\documentclass[lettersize,journal]{IEEEtran}
\usepackage[utf8]{inputenc}
\usepackage{dtk-logos} 
\usepackage{graphicx}
\usepackage{hyperref}
\usepackage{enumitem}
\usepackage{amsmath,amsfonts,amssymb,amsthm}
\usepackage[noadjust]{cite}

\hypersetup{colorlinks=true,allcolors=blue,linktocpage}
\usepackage{mathtools}
\usepackage{bm,upgreek}
\usepackage{nomencl}
\usepackage{multirow}
\usepackage{float}
\usepackage{subcaption}
\usepackage{array}
\usepackage{algorithm}
\usepackage{algpseudocode}
\usepackage{orcidlink}
\makenomenclature

\begin{document}

	\title{Comparing Z3 and A3 PKM Heads: Which Is Superior and Why?}
	\author{Hassen Nigatu \raisebox{0.5ex}{\orcidlink{0000-0002-1825-0097}}, Lu Guodong \raisebox{0.5ex}{\orcidlink{0000-0002-1825-0097}}, Huixu Dong \raisebox{0.5ex}{\orcidlink{0000-0002-1825-0097}} 
	\thanks{Hassen Nigatu and Huixu Dong are with (Grasp Lab of the Mechanical Engineering Department, Zhejiang University, Hangzhou, 310058) and (Robotics Research Center of Yuyao, Yuyao Technology Innovation Center, No. 479, Yuyao, Ningbo City, 315400), Zhejiang, China.  \\ \texttt{Email: \{hassen@ust.ac.kr, huixudong@zju.edu.cn \}}}
	\thanks{Lu Guodong is with (the the Mechanical Engineering Department, Zhejiang University, Hangzhou, 310058) and (Robotics Research Center of Yuyao, Yuyao Technology Innovation Center, No. 479, Yuyao, Ningbo City, 315400), Hangzhou, China.  }}
	
	\markboth{Journal of Robotics and Automation Letters,~Vol.~xx, No.~x, September~2024}%
	{Stiffness Comparison of 3-PRS and 3-RPS PMs within the Parasitic and Orientation Workspaces}      
	
	\maketitle
%
\begin{abstract}
This study presents a comparison between the Sprint Z3 and A3 head parallel kinematics machines, distinguished by their joint sequence. The analysis focuses on performance attributes critical for precision machining—specifically, parasitic motion, workspace capability, stiffness performance over the independent and parasitic spaces, and condition number distribution. Although these machines are extensively utilized in precision machining for the aerospace and automotive industries, a definitive superior choice has not been identified for machining large components. Moreover, the distribution of stiffness across the configuration of parasitic space has not previously been addressed for either mechanism. This research reveals that despite identical parameters used and exhibiting similar parasitic motions, the Sprint Z3 demonstrates superior stiffness, workspace volume, and condition number distribution. This performance advantage is attributed to variations in joint and link sequence, which enhance deflection resilience—crucial for manufacturing large-scale components. This also results in a higher condition number and a larger workspace. The result highlights the importance of design architecture in the efficacy of parallel kinematics machines and suggest that seemingly minor differences can have significant impacts. 
\end{abstract}

\section{Introduction} \label{sec:intro}
\begin{figure*}[h!]
	\centering
	\includegraphics[width=\textwidth]{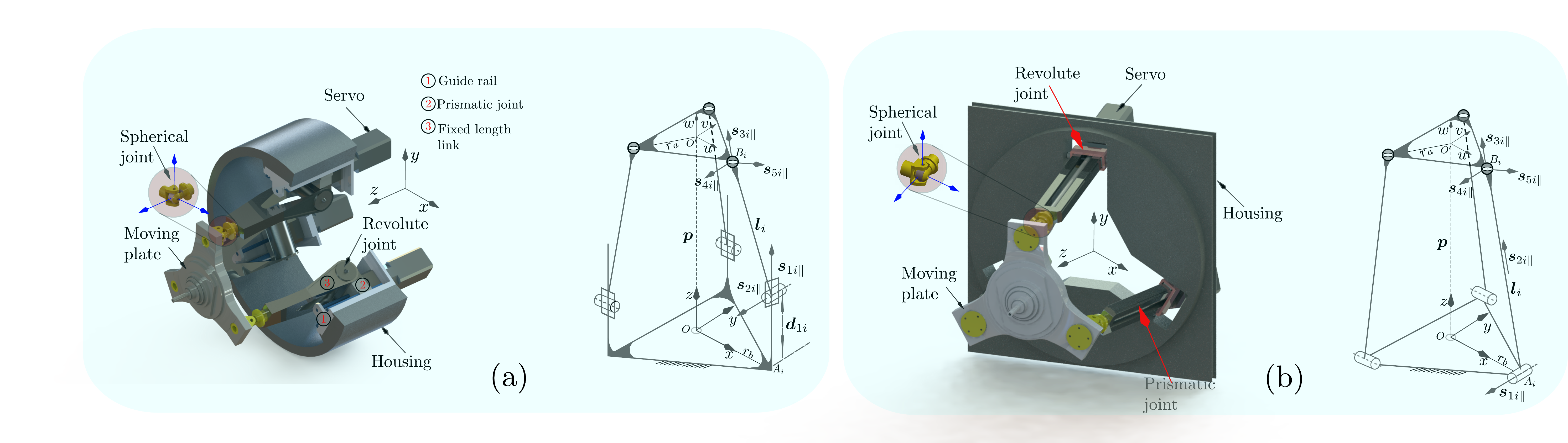} 
	\caption{3D model and schematics: (a) Sprint Z3 head (b) and A3 head }
	\label{fig:3d}
\end{figure*}

Parallel kinematic machines (PKMs) are favored in applications requiring high accuracy, rigidity, and superior dynamic performance \cite{inbook,starrag,pkmtricept}. Manipulators such as the Exechon \cite{ZHANG2016208}, Sprint Z3 \cite{WAHL2000}, and Tricept \cite{pkmtricept} have been effectively utilized in the automobile and aerospace industries for precision machining of large components. Moreover, the Delta robot has been widely used in 3D printing. However, despite tremendous research efforts, the industry demand for PKMs has been poor in the past several decades. One of the most impressively successful manipulators along this track is the Sprint Z3 (Fig. \ref{fig:3d}(a)) of the Starrag Group \cite{starrag}.  In a recent development, Orizonaero, a major supplier of aircraft equipment, reported that the manufacturer of Sprint Z3 (Starrag Group) received a large order from the US government in 2016 \cite{StarragG98}. This indicates a growing demand for these types of PKMs, leading to ongoing research activities focused on building machines with higher stiffness, lower kinematic errors, and superior dynamic performance compared to existing ones.  

The popular Sprint Z3 manipulator, with its three equispaced vertical linear drives, has a counterpart known as the A3 PKM (Fig. \ref{fig:3d}(b)), or A3 head, achieved by rearranging joint sequences in a limb \cite{LI20131577}. Despite numerous studies conducted on these closely similar manipulators \cite{LIAN2016190,YU2018137,0954406215586233,Nigatu2020,Nigatu2021,Nigatu2021_opt}, only a few symmetric comparisons have been carried out \cite{Chen2014}. Therefore, this paper addresses a comprehensive performance comparison of the two parallel kinematic machines (PKMs) based on their condition number distribution, workspace capability, parasitic motion, and stiffness, utilizing their Jacobian matrices. By setting all geometric parameters and stiffness coefficients, the evaluation is performed based on the constraint embedded Jacobian, which appears to differ due to variations in their joint sequence. The Jacobian is obtained using the analytic reciprocal screw method \cite{article,Nigatu2023,Nigatu2021_mmt,Nigatu2023ICCAS}. The novel dimensionally homogeneous Jacobian (DHJ) discussed in \cite{Nigatu2023,Nigatu2023ICCAS} is used to compute the condition number and workspace of the manipulators. This is because the conventional Jacobian is not suitable for evaluating the condition number and manipulability-based performance of this manipulator, due to its mixed linear and rotational freedoms \cite{Merlet2007,Pond2006,Nigatu2023,Nigatu2023ICCAS}. The stiffness modeling approach presented in \cite{ZHANG2016208,0954406215586233,LI20131577} with the analytic inverse Jacobian in \cite{Nigatu2021,Nigatu2023} is used to evaluate the stiffness performance. Parasitic motion governing equation at the velocity level is explicitly obtained from the constraint matrix as presented in \cite{Nigatu2021_mmt}. Hence, in this study, all performance evaluations rely on the velocity level relationship to ensure both consistency and simplicity, integrating motion and constraints seamlessly within the velocity framework.

The strange phenomenon known as parasitic motion has been identified in the Degree of Restriction (DoR) of the moving platform, significantly influencing the use of these Parallel Kinematic Machines (PKMs) for machining large components \cite{Ruiz2016,Nigatu2021_mmt}. This motion manifests as an undesired displacement or de-centering effect of the moving plate center along the restricted axis \cite{Huang2013,Carretero2000,Nigatu2021,Nigatu2021_mmt,Nigatu2021_opt,Nigatu2023}, which considerably affects the manipulator's accuracy. Furthermore, understanding the stiffness distribution along or around the direction of this motion is crucial. The stiffness performance within this parasitic configuration is obtained for the first time, whereas previous works typically evaluated stiffness within a workspace constructed by independent parameters. Therefore, this paper evaluates the complete stiffness performance of the Sprint Z3 and A3 head parallel kinematic machines, in addition to analyzing the condition number, workspace, and parasitic motion amplitude. Understanding these properties is essential for users to select machines best suited to their needs and for the development of improved design and control strategies to maximize machine potential in industrial applications.
                      
\subsection{Mechanisms description} \label{subsec:mech-description}
The two target mechanisms 3D model and schematics are shown in Fig. \ref{fig:3d}. Each limb of the manipulators comprises five one Degree of Freedom (DoF) joints, which are revolute, prismatic, and spherical, regardless of their order. Frame $O$ is attached at the center of the base plane, and Frame $O'$ is attached at the center of the moving platform. Lengths $r_b$ and $r_a$ represent the radii of the base and moving platform, respectively. $\boldsymbol{a}_i$ is the vector from $O'$ to center of spherical joint ($B_i$), based on frame $O$. Spherical joints located at $B_i$ connect the limb with the moving platform. 
All three limbs of both mechanisms are arranged to be located with a $120^\circ$ difference around the fixed frame $O$ to connect the two plates. The $x$ and $y$ axes of frame $O$ lie on the base plate, with $x$ directed along the first limb. When the isolated limb is detached from the top platform at the spherical joints, its end-effector is point $B_i$. However, the limb vectors are referenced in frame $O$. The position vectors $\boldsymbol{l}_{i}$ is directed along the link passing through the center of spherical joint. Note that the first (revolute) joint of the A3 PKM is fixed at the base, while the revolute joint of the Sprint Z3 PKM moves together with the actuated prismatic joint. Vector $\boldsymbol{p}$ represents the position of the center of the moving plate with respect to the fixed frame $O$. Vectors $\boldsymbol{s}_{ji\parallel}$ indicates the direction of $j^{th}$ joint in  the $i^{th}$ limb.
  
\section{Parasitic Motions and workspace analysis}
The inverse rate kinematics of two mechanisms can be described in a general form shown in Eq. (\ref{onee}).
\begin{equation}
	\begin{aligned}
		\boldsymbol{\dot{q}} =  \boldsymbol{G}^\top \boldsymbol{\mathcal{\dot{X}}} \\ 
		\begin{bmatrix}
			\dot{\boldsymbol{q}}_a \\ \dot{\boldsymbol{q}}_c
		\end{bmatrix}=
		\begin{bmatrix}
			\boldsymbol{G}_{av}^\top & \boldsymbol{G}_{a\omega}^\top\\ \boldsymbol{G}_{cv}^\top & \boldsymbol{G}_{c\omega}^\top 
		\end{bmatrix} \begin{bmatrix}  \boldsymbol{v} \\ \boldsymbol{\omega} \end{bmatrix}
		\label{onee}
	\end{aligned}
\end{equation}
where $\boldsymbol{\mathcal{\dot{X}}}$  is the constraint compatible task motion which comprise both the parasitic and independent velocity components.

\subsection{Detection and identification of parasitic motion}
As far as $\dot{\boldsymbol{q}}_c$ in Eq. (\ref{onee}) is a constraint, it should always be zero. Therefore, the active joint rate equation must satisfy the following condition.
\begin{equation}
\boldsymbol{G}_c^\top\boldsymbol{\mathcal{\dot{X}}}  = \boldsymbol{0}
\label{twoo}
\end{equation}
To meet the structural constraint outlined in Eq. (\ref{twoo}), $\boldsymbol{\mathcal{\dot{X}}} \in \mathbb{R}^6$ should correspond to a feasible velocity profile for the manipulator. Typically, any arbitrary user input $\boldsymbol{\mathcal{\dot{X}}}_a$ can be transformed into feasible motion by projecting this input onto the motion space, via the following relation.

\begin{equation}
	\begin{aligned}
		\boldsymbol{G}_c^\top\boldsymbol{\mathcal{\dot{X}}} &= \boldsymbol{0} \\
		 \boldsymbol{G}_c^\top \Big( \boldsymbol{I} - \boldsymbol{G}_c \boldsymbol{G}_c^{\dagger} \Big)\boldsymbol{\mathcal{\dot{X}}}_a  &= \boldsymbol{0}
	\end{aligned}	
	\label{eq:con_compa}
\end{equation}
In Eq. (\ref{eq:con_compa}), the function represented by $\boldsymbol{I} - \boldsymbol{G}_c \boldsymbol{G}_c^{\dagger}$ plays a crucial role in transforming arbitrary user inputs into motion that comply with the structural constraints of the manipulator, where $(\cdot^\dagger)$ denotes the Moore–Penrose inverse. It projects any spatial vector $\boldsymbol{\mathcal{\dot{X}}}_a$ onto the null space of $\boldsymbol{G}_c$. This projection matrix effectively filters or modifies the arbitrary user input, ensuring $\boldsymbol{\mathcal{\dot{X}}}$ remains within the permissible motion space.

With the assurance adhering to the above constraint policy, we need to obtain dependent or parasitic motions of end-effector as a function of independent ones. For this, the first step is to detect and identify the parasitic motion from the independent ones. For the manipulators discussed here, independent and parasitic motions are known. However, in other cases, the detection and identification of parasitic motions can be accomplished using the matrix $\boldsymbol{P}$.

\begin{equation}
	\begin{aligned}
		\boldsymbol{P} &= \boldsymbol{I} - \boldsymbol{G}_c \boldsymbol{G}_c^{\dagger} \\ 
		               &= \begin{bmatrix} \boldsymbol{p}_{ij\perp} & \boldsymbol{p}_{ij\parallel} \end{bmatrix} 
	\end{aligned}
\label{eq:three}
\end{equation}
where $\boldsymbol{p}_{ij\perp}^\top = \begin{bmatrix} p_{i1\perp} & p_{i2\perp} & p_{i3\perp} \end{bmatrix}^\top $ and  = $\boldsymbol{p}_{ij\parallel}^\top = \begin{bmatrix} p_{i1\parallel} & p_{i2\parallel} & p_{i3\parallel} \end{bmatrix}^\top $ for $i = 1, \dots, 6$ represents the basis vector of $\boldsymbol{P}$ for the $i^{\text{th}}$ row and $j^{\text{th}}$ column. The first vector term along the $i^{\text{th}}$ rows represents the term that filters the linear component of $\boldsymbol{\mathcal{\dot{X}}}_a$, while the second term filters out the constraint-violating rotational input of $\boldsymbol{\mathcal{\dot{X}}}_a$. For the Sprint Z3 and A3 heads analyzed in this study, the matrix $\boldsymbol{P}$ unveils two distinct forms across different configurations, namely:

\begin{itemize}
	\item \texttt{Home configuration}: This configuration maintains a constant orientation while allowing arbitrary changes in heave, i.e., the case where the moving plate and base plates are parallel.
	\item \texttt{Rotated configuration}: This configuration involves changes in orientation, either alone or in combination with heave changes. In this state, the entries of the matrix $\boldsymbol{P}$ also change, as they are dependent on the configuration.
\end{itemize}

In the \texttt{Home configuration}, all entries of $\boldsymbol{P}$ are zero except for $p_{33_{\perp}}$, $p_{41_{\perp}}$, and $p_{52{\parallel}}$ as shown in Eq. (\ref{eq:five}). These three terms correspond to $v_z$, $\omega_x$, and $\omega_y$, respectively.      

\begin{equation}
\boldsymbol{P}=
\begin{bmatrix}
0 & 0 & 0 & 0 & 0 & 0 \\ 0 & 0 & 0 & 0 & 0 & 0\\ 0 & 0 & 1 & 0 & 0 & 0 \\ 0 & 0 & 0 & 1 & 0 & 0 \\ 0 & 0 & 0 & 0 & 1 & 0 \\ 0 & 0 & 0 & 0 & 0 & 0
\end{bmatrix}
\label{eq:five}
\end{equation}

 At this configuration, any arbitrary non-zero $v_x$, $v_y$, and $\omega_z$ components of $\boldsymbol{\mathcal{\dot{X}}}_a$ gives $v_x=0$, $v_y=0$, and $\omega_z=0$ components of $\boldsymbol{\mathcal{\dot{X}}}$. Hence this input cannot generate motion to the manipulator. If these components were independent input, they would not be nullified by matrix $\boldsymbol{P}$. However, their nullification indicates that they are considered constraints. Conversely, if we input either $\omega_x$, $\omega_y$, or $v_z$ components of $\boldsymbol{\mathcal{\dot{X}}}_a$, or combine them simultaneously, the configuration of the manipulator changes, which also modifies the entries of $\boldsymbol{P}$. Thus, these terms can generate motion for the manipulator, enabling it to move from the home configuration to the desired pose. This action identifies them as independent components in $\boldsymbol{\mathcal{\dot{X}}}$, essential for achieving the target configuration. At this point, if the $v_x$, $v_y$, and $\omega_z$ components of $\boldsymbol{\mathcal{\dot{X}}}$ are non-zero, it indicates the presence of parasitic motion in the manipulator. Consequently, this motion has been detected automatically. It means, the moving plate has only three controllable variables $z$, $\theta$ and $\phi$ in the Cartesian space. These are one translation along $z$ and two rotations about $x$ and $y$ axes. Thus, with the help of Eqs. (\ref{eq:three}, \ref{eq:five}), the detection and identification is achieved. 
 \begin{equation} \boldsymbol{\mathcal{\dot{X}}}_a   = 
\begin{bmatrix}
0&0& \dot{z}& {\omega_x}&{\omega_y}&0
\end{bmatrix}^\top
\label{eq:six}
\end{equation} 

In Eq. (\ref{eq:six}), the user can specify only the $v_z$, $\omega_x$, and $\omega_y$ components of $\boldsymbol{\mathcal{\dot{X}}}_a$. Equation (\ref{eq:three}) then automatically determines the remaining components of $\boldsymbol{\mathcal{\dot{X}}}_a$, as these values cannot be arbitrary. Despite parasitic motion is being undesirable from the user's or task's perspective, it is essential for the manipulator to satisfy the constraints. Upon isolating the parasitic terms from the independent ones with Eq. (\ref{eq:con_compa}), it becomes necessary to define a coupling relationship between parasitic and independent terms. This can be done by rearranging the components from Eq. (\ref{eq:con_compa}). The first step toward this is obtaining the motion and constraint wrench for the inverse Jacobian as follows:

\begin{subequations}
	\begin{equation}
		\boldsymbol{G}_{PRS}^\top = \begin{bmatrix}
			\boldsymbol{w}_{ai}^\top \\ \boldsymbol{w}_{ci}^\top 
		\end{bmatrix}=
		\begin{bmatrix}
			\dfrac{\boldsymbol{l}_{1i}^\top}{{\boldsymbol{l}_{1i}^\top s_{1i\parallel}}} & \dfrac{(\boldsymbol{l}_{1i}\times \boldsymbol{a}_i)\top}{{{\boldsymbol{l}_{1i}^\top s_{1i\parallel}}}}  \\
			\boldsymbol{s}_{2i{\parallel}}^\top & (\boldsymbol{s}_{2i{\parallel}}\times \boldsymbol{a}_i)^\top \\
		\end{bmatrix}  \label{eq:Gprs}
	\end{equation}
	
	\begin{equation}
		\boldsymbol{G}_{RPS}^\top  =  \begin{bmatrix}
			\boldsymbol{w}_{ai}^\top \\ \boldsymbol{w}_{ci}^\top 
		\end{bmatrix}=
		\begin{bmatrix} 
			\dfrac{\boldsymbol{l}_{1i}^\top}{{{\boldsymbol{l}_{1i}^\top s_{2i\parallel}}}} & \dfrac{(\boldsymbol{l}_{1i}\times \boldsymbol{a}_i)^\top}{{{{\boldsymbol{l}_{1i}^\top s_{2i\parallel}}}}}  \\
			s_{1i{\parallel}}^\top & (s_{1i{\parallel}}\times \boldsymbol{a}_i)^\top \\
		\end{bmatrix} \label{eq:Grps}
	\end{equation} 
\end{subequations}

where, $\boldsymbol{w}_{ai}$ and $\boldsymbol{w}_{ci}$ represent the active and constraint wrenches of the three limbs, respectively, accompanied by their corresponding direction and moment vectors. Furthermore, $\boldsymbol{a}_i$ denotes the position vector of the moving plate, extending from the center of the spherical joint to the center of the moving plate.
Matrices $\boldsymbol{G}_{PRS}^\top$ and $\boldsymbol{G}_{RPS}^\top$ are the inverse Jacobian of Z3 and A3 PKMs, respectively. To explicitly define dependent motions in terms of independent ones, we must consider the direction and moment vectors of the constraint wrenches obtained in Eq. (\ref{eq:Gprs}) and (\ref{eq:Grps}) on an element-by-element basis as follows.
 
\begin{equation}
	\begin{aligned}
		\boldsymbol{G}_{ci}^\top \boldsymbol{\mathcal{\dot{X}}} &= 0 \\ 
		\begin{bmatrix} \boldsymbol{s}_{ij\parallel}^\top  & (\boldsymbol{s}_{ij\parallel} \times \boldsymbol{a}_i)^\top \end{bmatrix} \begin{bmatrix} \boldsymbol{v} \\ \boldsymbol{\omega} \end{bmatrix} &= 0 \\
		\boldsymbol{s}_{ij\parallel}^\top\boldsymbol{v} + [\boldsymbol{s}_{ij\parallel}]_\times \boldsymbol{a}_i\boldsymbol{\omega} & = 0
	\end{aligned}	\label{eq:con_ith}
\end{equation}
 
 The vector $ \boldsymbol{s}_{ij\parallel} = \begin{bmatrix} -s\xi_i & c\xi_i &0 \end{bmatrix}^\top $ represents the direction of the constraint wrench, aligned parallel to the revolute joints of both manipulators, as the motion of the limbs is restricted by these joints. Unpacking Eq. (\ref{eq:con_ith}), we get  
 
 \begin{equation}
 	\begin{aligned}
 	-v_x s\xi_i + v_y c\xi_i + a_{iz}c\xi_i \omega_x + a_{iz}s\xi_i \omega_y + \\ ( -a_{ix}c\xi_i - a_{iy}s\xi_i)\omega_z = 0 
    \end{aligned} \label{eq:con_ith_exp}
 \end{equation}  

Collecting $v_x$, $v_y$ and $\omega_z$ terms and their coefficients in Eq. (\ref{eq:con_ith_exp}) to one side leads us to 

\begin{equation}
	\begin{aligned}
		\begin{bmatrix} -s\xi_i & c\xi_i (-a_{ix} c\xi_i - a_{iy} s\xi_i ) \end{bmatrix} \begin{bmatrix} v_x \\ v_y \\ \omega_z\end{bmatrix} =  \\  \begin{bmatrix}  -a_{iz}c\xi_i & -a_{iz} s\xi_i \end{bmatrix}  \begin{bmatrix} \omega_x \\ \omega_y \end{bmatrix} \\ 
	\end{aligned} \label{eq:con_ith_collected}
\end{equation}

Writing Eq. (\ref{eq:con_ith_collected}) for all three limbs, we get

\begin{equation}
	\begin{aligned}
		\boldsymbol{C}_1 \dot{\boldsymbol{x}}_d = \boldsymbol{C}_2 \dot{\boldsymbol{x}}_i \\ 
    \end{aligned} \label{eq:coupling}
\end{equation}

where $\boldsymbol{C}_1 = \begin{bmatrix} -s\xi_1 & c\xi_1 (-a_{1x} c\xi_1 - a_{1y} s\xi_1 ) \\  -s\xi_2 & c\xi_2 (-a_{2x} c\xi_2 - a_{2y} s\xi_2)  \\ -s\xi_3 & c\xi_3 (-a_{3x} c\xi_3 - a_{3y} s\xi_3 ) \end{bmatrix}$, $\dot{\boldsymbol{x}}_d = \begin{bmatrix} v_x \\ v_y \\ \omega_z\end{bmatrix} $, $\boldsymbol{C}_2 = \begin{bmatrix}  -a_{1z}c\xi_1 & -a_{1z} s\xi_1 \\  -a_{2z}c\xi_2 & -a_{2z} s\xi_2 \\ -a_{3z}c\xi_3 & -a_{3z} s\xi_3 \end{bmatrix} $ and $ \dot{\boldsymbol{x}}_i =  \begin{bmatrix} \omega_x \\ \omega_y \end{bmatrix}$ 
   
\vspace{2mm}
From Eq. (\ref{eq:coupling}), it is evident that $v_z$ is not related to the generation of parasitic motion, which is consistent with the detection and identification step, as well as with previous literature. Finally, by inverting $\boldsymbol{C}_1$, we obtain the coupling relation between parasitic and independent motions, as shown below.

\begin{equation}
	\begin{bmatrix}  v_x \\ v_y \\ \omega_z \end{bmatrix} =  \boldsymbol{C}_1^{-1}\boldsymbol{C}_2 \begin{bmatrix} \omega_x \\ \omega_y \end{bmatrix} = \boldsymbol{C} \begin{bmatrix}
		\omega_x \\ \omega_y
	\end{bmatrix}
	\label{eq:7}
\end{equation}

\subsection{Workspace and condition number evaluation}

The workspace analysis of these manipulators is based on the condition number of the Jacobian matrix. However, the condition numbers given in Eq. \ref{eq:Gprs} and Eq. \ref{eq:Grps} cannot be directly used, as the manipulator has mixed translational and rotational degrees of freedom (DoFs). Therefore, a dimensionally homogeneous Jacobian is required. We use the dimensionally homogeneous Jacobian proposed in the author's previous work \cite{Nigatu2023ICCAS} and \cite{Nigatu2023}. Consequently, the condition number is consistent and reflects the true performance of the manipulator, making it suitable for comparative analysis.


\section{Stiffness modeling and analysis}

It is crucial to understand that the constraint space also has important roles in stiffness of the structure as it does in the kinematics. These manipulators cannot generate forces in all six degrees of freedom ($\mathbb{R}^6$), leading to constraints that, while unable to produce work, must be included in stiffness analysis. Therefore, this section develops the stiffness models for the Sprint Z3 and A3 heads, taking into account all significant component compliances and structural constraints. 

For simplicity,  the following assumptions hold in this paper. 
\texttt{Assumption one:} Fixed and moving plates are rigid. Other parts are regarded as a series of spring systems. Therefore, the axial and bending compliance of joints, limbs, lead screws and their associated elements are taken into account as a compliant element. 

 It's important to note that this is a static analysis, which disregards any inertia-induced bending in the direction of freedom.
Considering each limb components as a series of connected springs, the moving plate deflection can be expressed with the following relation.

\begin{equation}
\delta\boldsymbol{x} =\sum_{j_a=1}^{f}\$_{ai}\delta{q}_{ai} +\sum_{j_c=1}^{6-f}\$_{ci}\delta{q}_{ci}
\label{one}
\end{equation}
In Eq. (\ref{one}), $\delta{q}_{ai}$ and $\delta{q}_{ci}$ are the intensity of corresponding joint twist or can be called the linear/angular deflection values in the respective directions. Twists $\$_{ai}$ and $\$_{ci}$ correspond to the elastic deflection screws. This relation implies the joint and moving plate infinitesimal displacements can be mapped by using the limb Jacobian. Consequently, their inverse relation is obtained as: 

  \begin{equation}
  	\begin{aligned}
  		\delta{\boldsymbol{q}}                                                                                    &= \boldsymbol{G}^\top \delta{\boldsymbol{\mathcal{X}}} \\ 
  		 \begin{bmatrix} \delta{\boldsymbol{q}}_a \\ \delta{\boldsymbol{q}}_c \end{bmatrix}                       &= \begin{bmatrix}  \boldsymbol{G}_{av}^{\top} & \boldsymbol{G}_{aw}^{\top} \\  \boldsymbol{G}_{aw}^{\top} & \boldsymbol{G}_{cw}^{\top} \end{bmatrix} \begin{bmatrix} \delta\boldsymbol{r} \\ \delta\boldsymbol{\alpha} \end{bmatrix} 
  	\end{aligned}  \label{eq:stif_irk}
  \end{equation}


If $\boldsymbol{\tau}=\begin{bmatrix}
\boldsymbol{f}^\top&\boldsymbol{m}^\top
\end{bmatrix}^\top$ is applied on the moving platform and $\delta\boldsymbol{x}=\begin{bmatrix}
\delta\boldsymbol{r}^\top & \delta \boldsymbol{\alpha}^\top
\end{bmatrix}^\top$ be deflection induced by $\boldsymbol{\tau}$ where, $\delta\boldsymbol{r}$ and  $\delta \boldsymbol{\alpha}$ are linear and angular deflections. The manipulator in a state of equilibrium under the action of external forces (and moments),$\boldsymbol{\tau}$, and internal reaction forces $\boldsymbol{w}$, the net work done on the system should be zero for any small, virtual displacement. This condition ensures that the system's internal energy is conserved, and it does not spontaneously move or deform further, which would imply an imbalance. This phenomenon is described by the virtual work principle (vwp) as follows.

 \begin{equation}
 	\delta{\boldsymbol{\mathcal{X}}^\top \boldsymbol{\tau} = \delta{\boldsymbol{q}}^\top \boldsymbol{w} \label{eq:vrp_generic}
} \end{equation}

where $ \boldsymbol{w} $ is the reaction force which counteract the applied external wrench to maintain equilibrium. $ \delta{\boldsymbol{\mathcal{X}}} $ and $ \delta{\boldsymbol{q}} $ are infinitesimally small displacements produces by $ \boldsymbol{\tau} $ and $ \boldsymbol{w} $ that do not alter the external forces acting on the system. From Eq.( \ref{eq:Gprs}) and Eq. (\ref{eq:Grps}), we derive Jacobian matrices that maps those deflections, encompassing the constraints. Hence, Eq. (\ref{eq:vrp_generic}) can be rewrite as 

\begin{equation}
	\begin{aligned}
		\delta{\boldsymbol{\mathcal{X}}}^\top \boldsymbol{\tau} = \delta{\boldsymbol{\mathcal{X}}}^\top \boldsymbol{G}   \boldsymbol{w} \\ 
	\end{aligned} \label{eq:vrp_gen_exp}
	\end{equation}

\texttt{Assumption two:} In the analysis, we model the individual components within a limb as a series of linearly connected elements. Adhering to Hooke's Law, which posits that the force required to extend or compress a spring by a certain distance is directly proportional to that distance, we formalize this relationship as:

\begin{equation}
	\boldsymbol{w} = \boldsymbol{\mathcal{K}} \delta{\boldsymbol{q}} \label{eq:hooks_law}
\end{equation}
where $\boldsymbol{\mathcal{K}} \in \mathbb{R}^{6 \times 6}$ is the actuation and constraint stiffness matrix, $\boldsymbol{w} \in \mathbb{R}^6$ is the force applied, and $\delta{\boldsymbol{q}} \in \mathbb{R}^6$ is the displacement caused by the force. This formulation underpins our assumption of linear elasticity for the components, facilitating the analysis of their response to applied forces.  

Reorganizing Eq. (\ref{eq:vrp_gen_exp}) we get 

\begin{equation}
	\boldsymbol{\tau} = \boldsymbol{G}  \boldsymbol{w}  \label{eq:tau_init}
\end{equation}

Then, substitute Eq. (\ref{eq:hooks_law}) and Eq. (\ref{eq:stif_irk}) into Eq. (\ref{eq:tau_init}), we get 

\begin{equation}
	\begin{aligned}
		\boldsymbol{\tau} &= \boldsymbol{G} \boldsymbol{\mathcal{K}} \delta{\boldsymbol{q}} \\ 
		                  &= \boldsymbol{G} \boldsymbol{\mathcal{K}} \boldsymbol{G}^\top \delta{\boldsymbol{\mathcal{X}}} \\ 
		                  &= \boldsymbol{K}\delta{\boldsymbol{\mathcal{X}}}
	\end{aligned} \label{eq:tau}
\end{equation}

Equation (\ref{eq:tau}) can be further expanded for unveiling the details as 

\begin{equation}
	\begin{bmatrix} \boldsymbol{\tau}_a \\ \boldsymbol{\tau}_c	\end{bmatrix}  = \begin{bmatrix} \boldsymbol{K}_{a\parallel} & \boldsymbol{K}_{a\perp}	 \\ \boldsymbol{K}_{c\parallel} & \boldsymbol{K}_{c\perp} \end{bmatrix} \begin{bmatrix} \delta{\boldsymbol{r}} \\ \delta{\boldsymbol{\alpha}}	\end{bmatrix} 
\end{equation}

\subsection{Formulation of stiffness matrix}

\begin{figure}[!ht]
	\centering
	\includegraphics[width=\columnwidth]{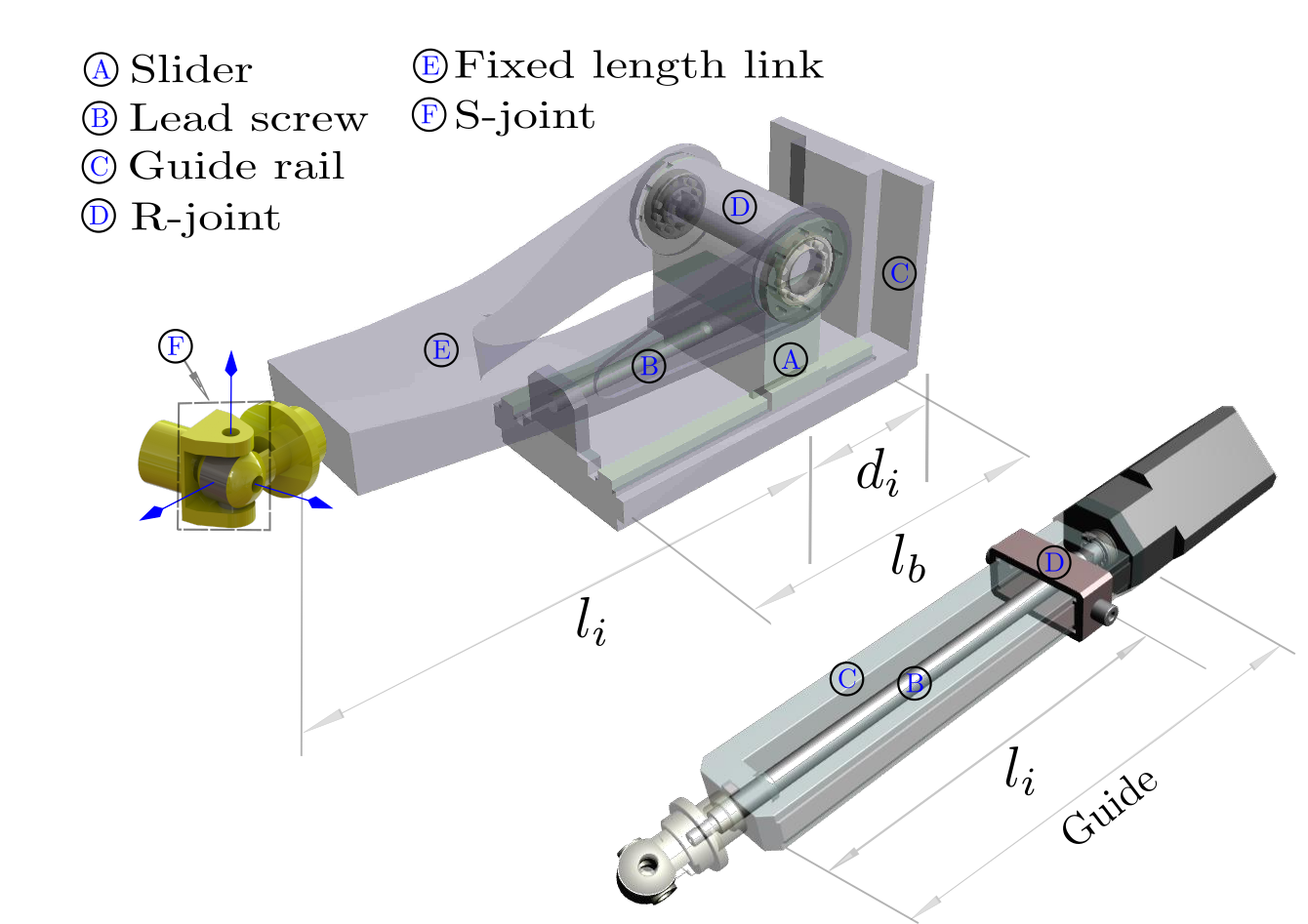}
	\caption{Limb of the manipulators. a ) A3 head  b) Sprint Z3}
	\label{limb}
\end{figure}

The formulation of the stiffness matrix $\boldsymbol{\mathcal{K}}$ involves calculating the axial and torsional stiffness of the components related to actuation and constraints, and compiling their elasticity information into matrix form. For convenience and for the purpose of comparison, the stiffness coefficient of the limb components of the two manipulator shown in FIg. \ref{limb} is set at 1e6.

The axial stiffness for each limb is calculated using the equation:

\begin{equation}
	\dfrac{1}{k_{ai}}  = \dfrac{1}{k_{cai}} + \dfrac{1}{k_{rji}} + \dfrac{1}{k_{lbi}}
\end{equation}

where:
\begin{itemize}
	\item $k_{cai}, k_{rji},$ and $k_{lbi}$ denote the axial stiffness coefficients of the Carriage assembly, Revolute joints, and Limb body, respectively.
	\item $k_{ai}$ represents the overall axial stiffness coefficient for limb $i$.
\end{itemize}

The torsional stiffness coefficients at the distal point of each limb are measured along the axis of the revolute joint. This axis is parallel to $\boldsymbol{s}_{1i\parallel}$ for the A3 head and $\boldsymbol{s}_{2i\parallel}$ for the Sprint Z3 head, respectively. 
Then, the torsional stiffness coefficient $k_{ci}$ can be determined by calculating the inverse sum of the torsional stiffness coefficients of the spherical joint and the limb body assembly. 

\begin{equation}
	\dfrac{1}{k_{ci}} = 	\dfrac{1}{k_{si}} + 	\dfrac{1}{k_{lbi}}
\end{equation}
where $k_{si}$ is a configuration dependent stiffness coefficients of the spherical joint along and about the corresponding axes while limb body stiffness coefficient $k_{lbi}$ can be obtained via FEM analysis in realistic case. The derivation of  $k_{si}$ is given as follows. 

\begin{equation}
	\boldsymbol{R}_\mathcal{F}^\top  \boldsymbol{K}_s \boldsymbol{R}_\mathcal{F}  
\end{equation}

where $\boldsymbol{R}_\mathcal{F} = \begin{bmatrix} \boldsymbol{x} & \boldsymbol{y} & \boldsymbol{z}	\end{bmatrix} $ is the rotation matrix of spherical joint frame with respect the fixed frame and $\boldsymbol{K}_s = \begin{bmatrix}  k_{six} & 0 & 0 \\ 0 &  k_{siy} & 0 \\ 0 & 0 &  k_{siz}	\end{bmatrix}   $ is a set of stiffness coefficients for the spherical joints correspond the three axes. The orientation matrix can $\boldsymbol{R}_\mathcal{F}$ can be obtained from the following relationship: 

\begin{equation} \begin{aligned}
		\boldsymbol{H}_p &= \boldsymbol{H}\boldsymbol{R}_z(\xi_i)\boldsymbol{H}\boldsymbol{t}_x(r_b)\boldsymbol{H}\boldsymbol{t}_z(d_i)\boldsymbol{H}\boldsymbol{R}_y(\theta_{2i})\boldsymbol{H}\boldsymbol{T}_z(l_i)\times \\& ~~~~\boldsymbol{H}\boldsymbol{R}_y(\theta_{3i})\boldsymbol{H}\boldsymbol{R}_x(\theta_{4i})\boldsymbol{H}\boldsymbol{R}_z(\theta_{5i})   \end{aligned} 
 \label{eq:limb_trans}
\end{equation}
where \ref{eq:limb_trans}, $\boldsymbol{H}_p = \begin{bmatrix} \boldsymbol{R} & \boldsymbol{p} \\ \boldsymbol{0} & 1 \end{bmatrix}  $ represents the pose of the moving platform. Then, the spherical joint angles are simply obtained by extracting the rotational terms and equating the left-hand side and right-hand side of $\boldsymbol{H}{pr}^{-1}\boldsymbol{H}_p = \boldsymbol{H}_s$. Then, $\boldsymbol{R}_\mathcal{F}$ is established out of it.  

This approach effectively combines the contributions of both elements to the overall torsional stiffness of each limb.

The comprehensive stiffness matrix $\boldsymbol{\mathcal{K}}$ is an upper diagonal matrix formulated by assembling the stiffness contributions from all limb components into a diagonal matrix $\boldsymbol{\mathcal{K}}_{ac}$ and integrating coordinate transformation and Jacobian matrices adjustments:

\begin{equation}
	\begin{aligned}
		\boldsymbol{K} &= \boldsymbol{G} \boldsymbol{\mathcal{K}}\boldsymbol{G}^\top \\
		&= \begin{bmatrix} \boldsymbol{G}_a & \boldsymbol{G}_c \end{bmatrix}  \begin{bmatrix}  \boldsymbol{\mathcal{K}}_a & \boldsymbol{0} \\ \boldsymbol{0} & \boldsymbol{\mathcal{K}}_c \end{bmatrix}  \begin{bmatrix}  \boldsymbol{G}_a^\top  \\ \boldsymbol{G}_c^\top \end{bmatrix} 
	\end{aligned}
\end{equation}
where $\boldsymbol{G}$ embodies the combined actuation ($\boldsymbol{G}_a$) and constraint ($\boldsymbol{G}_c$) Jacobian matrices, and if the target coordinate is not the center of the moving platform, it can be adjusted through the $ 6 \times 6$ adjoint transformation matrix. 

Furthermore, the bending stiffness is included in $\boldsymbol{\mathcal{K}}$ by evaluating the reciprocal sum of the stiffness coefficients for each component, ensuring a comprehensive representation of the manipulator's elastic response to external forces and maintaining structural integrity under load.

\subsubsection{Simulation parameters}

Both manipulators are designed to have the same size for appropriate comparison and these parameters are given in Table \ref{tab:geometric_parameters_prs}. 

\begin{table}[!htb]
	\centering
	\caption{Geometric Parameters of the PRS Mechanism}
	\label{tab:geometric_parameters_prs}
	\begin{tabular}{ccc}
		\hline
		{Parameter} & {Description} & {Value} \\
		\hline
		$r_a$ & Radius of the base & 350 mm \\
		$r_b$ & Radius of moving plate & 250.00 mm \\
		$l$ & Link length & 642.3 mm \\
		\hline
	\end{tabular}
\end{table}

%

In the case of the A3 head, where $l$ represents the prismatic joint length, this length will be extended and shortened based on the orientation. Hence, the prismatic joint lengths $d_i$ for the Sprint Z3 and $l_i$ for the A3 head can respectively be obtained as follows:

 \begin{equation}
 	\begin{aligned}
 		d_i =g_{iz} -  \sqrt{ l^2 - g_{ix}} ~\text{for the Sprint Z3}\\
 		l_i = \sqrt{g_{ix}^2 + g_{iz}^2} ~\text{for the A3 head}\\
 	\end{aligned} 	
 \end{equation} 

where $\boldsymbol{g}_i = \begin{bmatrix} g_{ix} & g_{iy} & g_{iz} \end{bmatrix}^ \top = \boldsymbol{R}_z(\xi_i)^\top (\boldsymbol{p} + \boldsymbol{a}_i) - \boldsymbol{T}_x(r_b) $.


%
%
%
%
%
%

\section{Result discussion}
\begin{figure}[!htb]
 	\centering
 	\begin{subfigure}{0.5\columnwidth}
 		\centering
 		\includegraphics[width=\textwidth]{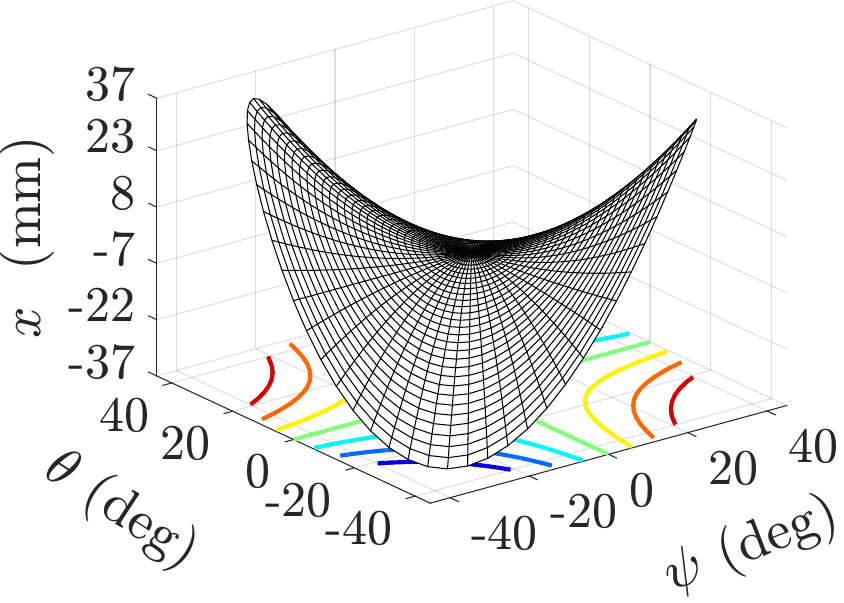}
 		\caption{}\label{fig:x_par}
 	\end{subfigure}\hfill
 	\begin{subfigure}{0.5\columnwidth}
 		\centering
 		\includegraphics[width=\textwidth]{x_par}
 		\caption{}\label{fig:x_par_A3}
 	\end{subfigure}\hfill
 	\caption{The $x$ translational parasitic motion. (a) Sprint Z3. (b)  A3 head}
 	\label{fig:x_pa}
 \end{figure}

\begin{figure}[!htb]
	\centering
	\begin{subfigure}{0.5\columnwidth}
		\centering
		\includegraphics[width=\textwidth]{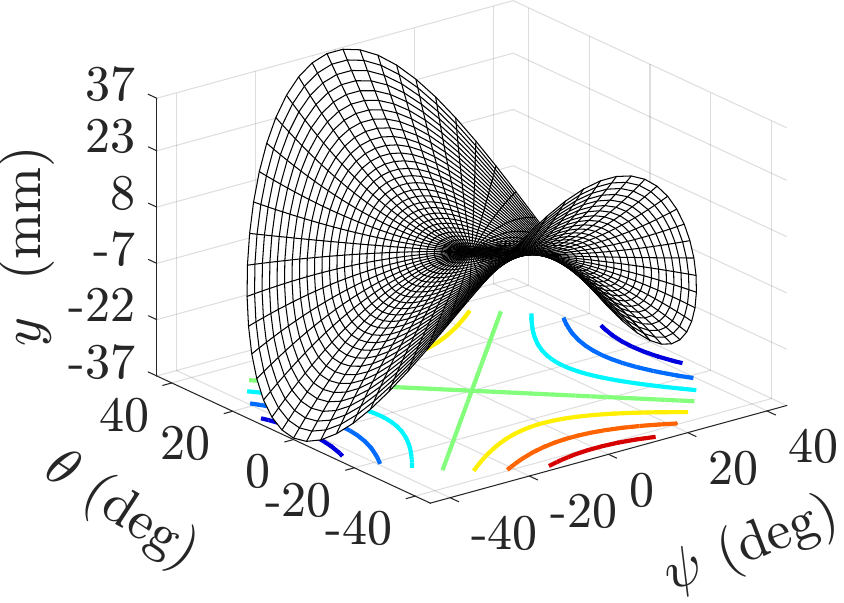}
		\caption{}\label{fig:y_par}
	\end{subfigure}\hfill
	\begin{subfigure}{0.5\columnwidth}
		\centering
		\includegraphics[width=\textwidth]{y_par}
		\caption{}\label{fig:y_par_A3}
	\end{subfigure}\hfill
	\caption{The $y$ translational parasitic motion.  (a) Sprint Z3. (b)  A3 head}
	\label{fig:y_pa}
\end{figure}

\begin{figure}[!htb]
	\centering
	\begin{subfigure}{0.5\columnwidth}
		\centering
		\includegraphics[width=\textwidth]{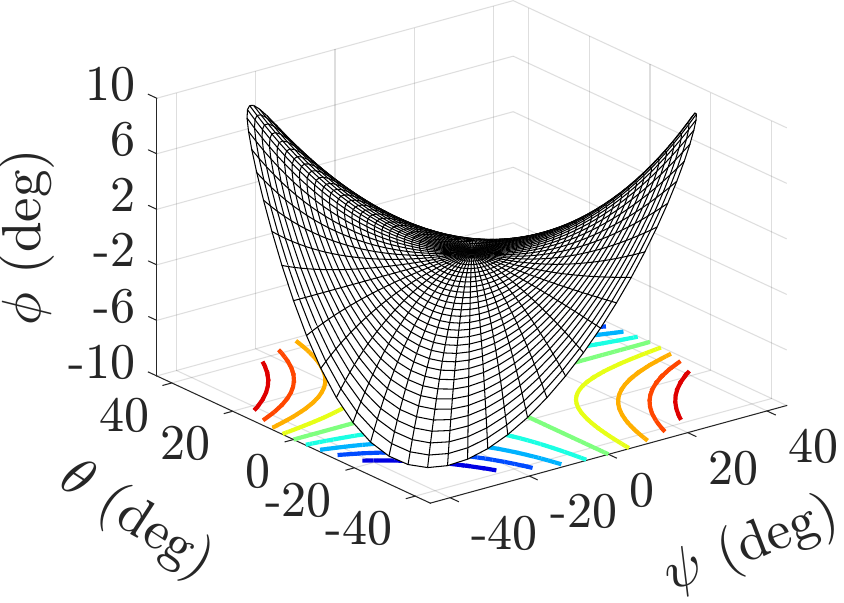}
		\caption{}\label{fig:z_par}
	\end{subfigure}\hfill
	\begin{subfigure}{0.5\columnwidth}
		\centering
		\includegraphics[width=\textwidth]{z_par}
		\caption{}\label{fig:z_par_A3}
	\end{subfigure}\hfill
	\caption{The $z$ rotational parasitic motion.  (a) Sprint Z3. (b)  A3 head}
	\label{fig:z_pa}
\end{figure}

\begin{figure}[!htb]
	\centering
	\begin{subfigure}{0.5\columnwidth}
		\centering
		\includegraphics[width=\textwidth]{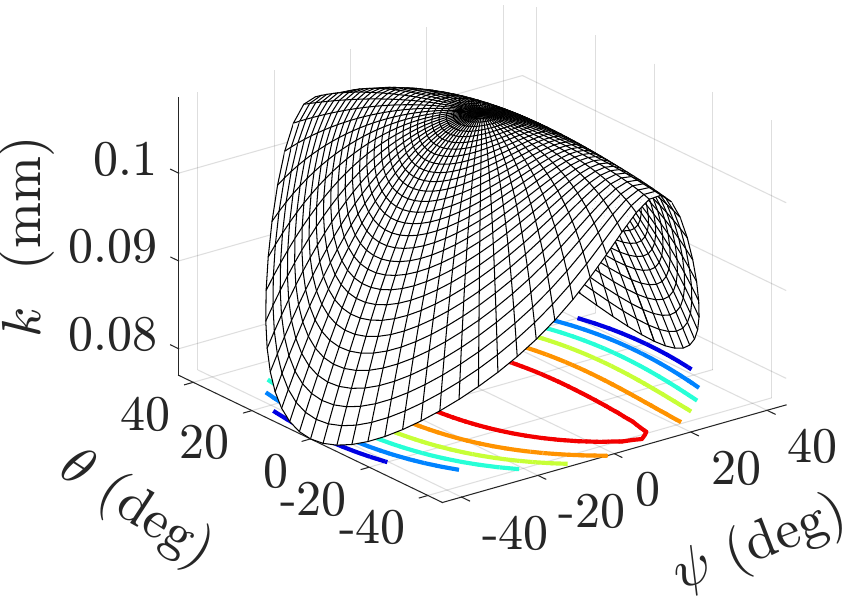}
		\caption{}\label{fig:cond}
	\end{subfigure}\hfill
	\begin{subfigure}{0.5\columnwidth}
		\centering
		\includegraphics[width=\textwidth]{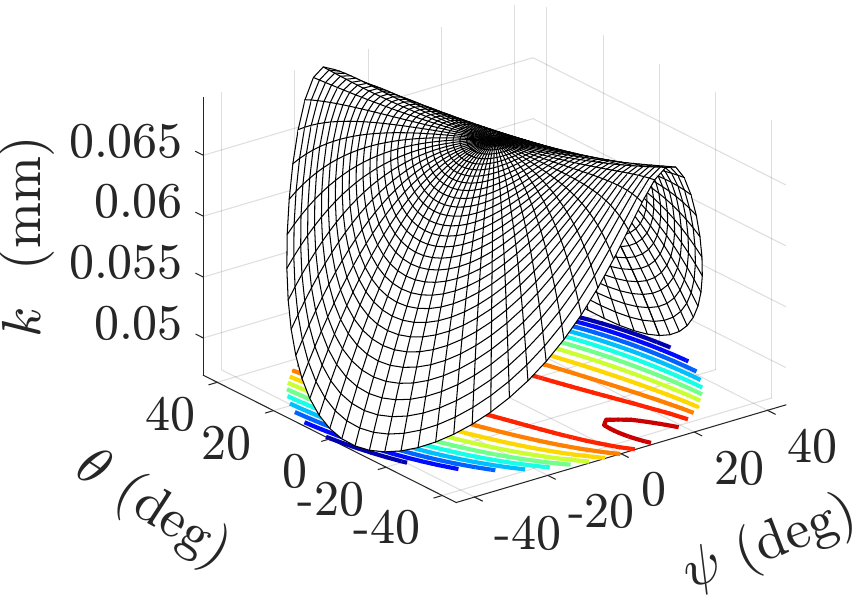}
		\caption{}\label{fig:cond_A3}
	\end{subfigure}\hfill
	\caption{Condition number ($k$).  (a) Sprint Z3. (b)  A3 head}
	\label{fig:cond_no}
\end{figure}

\begin{figure}[!htb]
	\centering
	\begin{subfigure}{0.5\columnwidth}
		\centering
		\includegraphics[width=\textwidth]{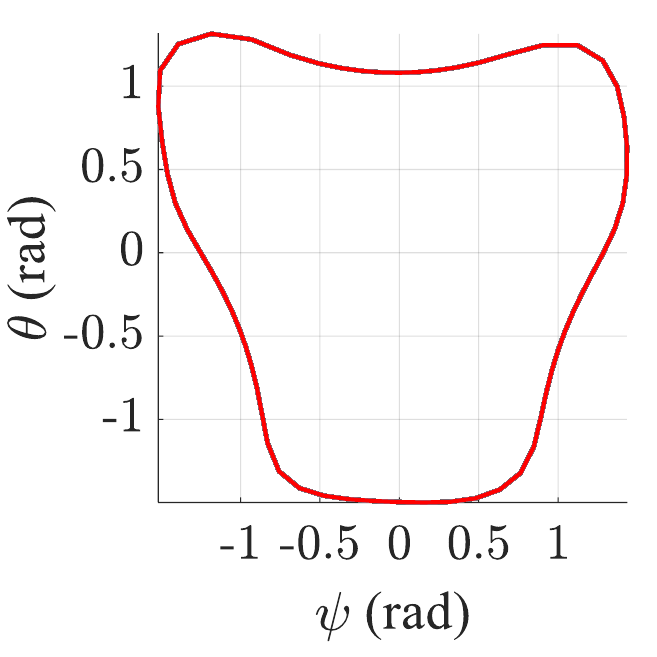}
		\caption{}\label{fig:ws_prs}
	\end{subfigure}\hfill
	\begin{subfigure}{0.5\columnwidth}
		\centering
		\includegraphics[width=\textwidth]{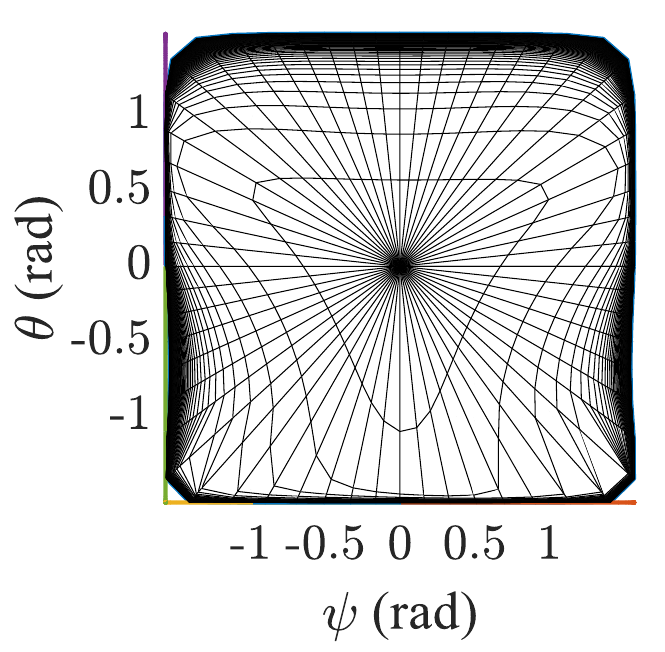}
		\caption{}\label{fig:ws_rps}
	\end{subfigure}\hfill
	\caption{Workspace.  (a) Sprint Z3. (b)  A3 head}
	\label{fig:workspace}
\end{figure}

\begin{figure}[!htb]
	\centering
	\begin{subfigure}{0.5\columnwidth}
		\centering
		\includegraphics[width=\textwidth]{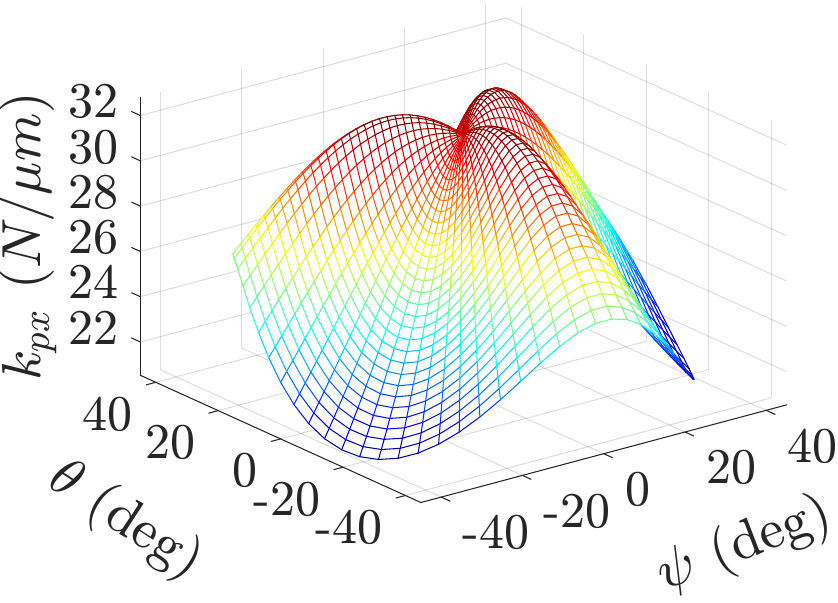}
		\caption{}\label{fig:kpx_A3}
	\end{subfigure}\hfill
	\begin{subfigure}{0.5\columnwidth}
		\centering
		\includegraphics[width=\textwidth]{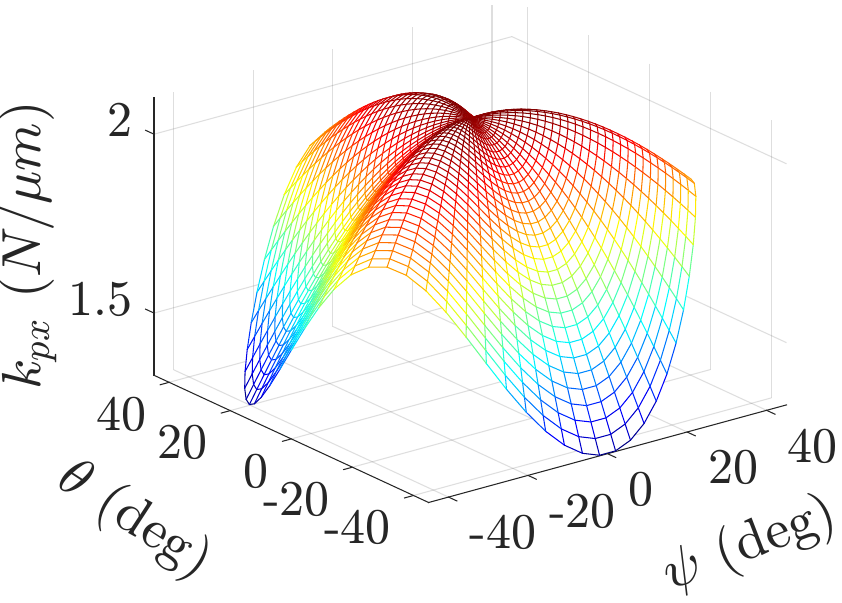}
		\caption{}\label{fig:kpx_z3}
	\end{subfigure}\hfill
	\caption{The x-axis axial stiffness distribution over rotational workspace.  (a) Sprint Z3. (b)  A3 head}
	\label{fig:kpx}
\end{figure}

\begin{figure}[!htb]
	\centering
	\begin{subfigure}{0.5\columnwidth}
		\centering
		\includegraphics[width=\textwidth]{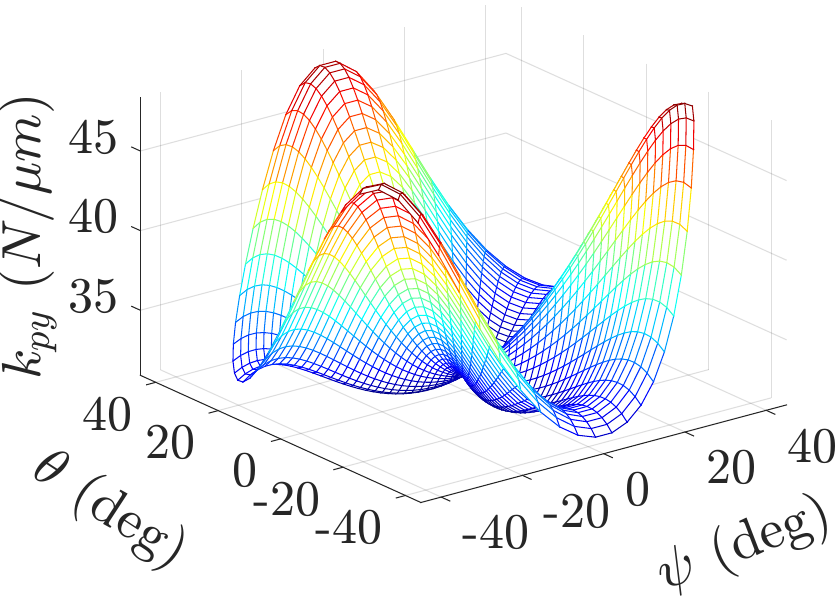}
		\caption{}\label{fig:kpy_A3}
	\end{subfigure}\hfill
	\begin{subfigure}{0.5\columnwidth}
		\centering
		\includegraphics[width=\textwidth]{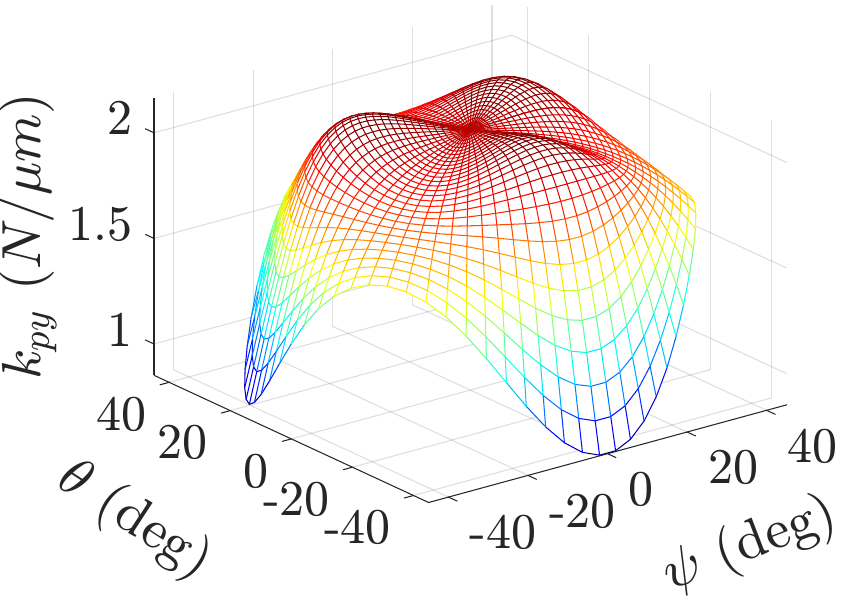}
		\caption{}\label{fig:kpy_z3}
	\end{subfigure}\hfill
	\caption{The y-axis axial stiffness distribution over rotational workspace.  (a) Sprint Z3. (b)  A3 head}
	\label{fig:kpy}
\end{figure}

\begin{figure}[!htb]
	\centering
	\begin{subfigure}{0.5\columnwidth}
		\centering
		\includegraphics[width=\textwidth]{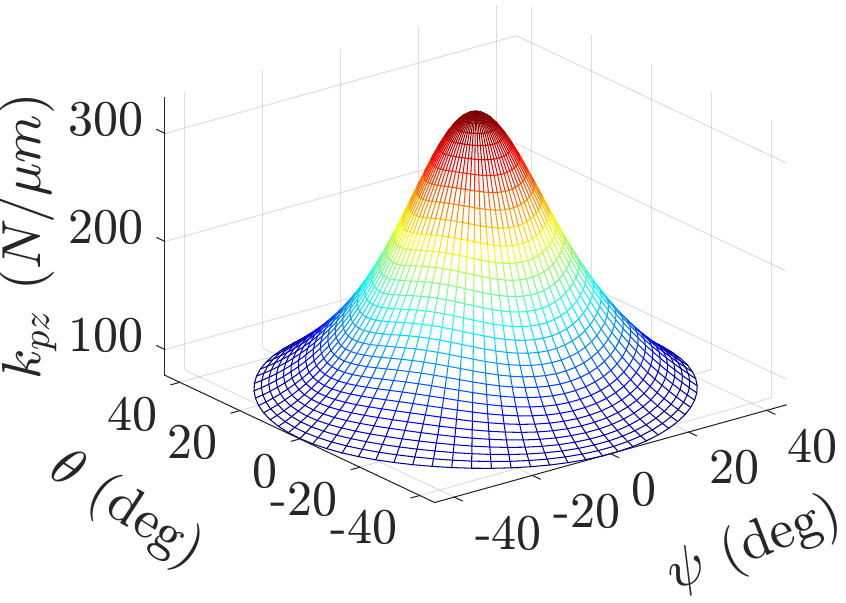}
		\caption{}\label{fig:kpz_A3}
	\end{subfigure}\hfill
	\begin{subfigure}{0.5\columnwidth}
		\centering
		\includegraphics[width=\textwidth]{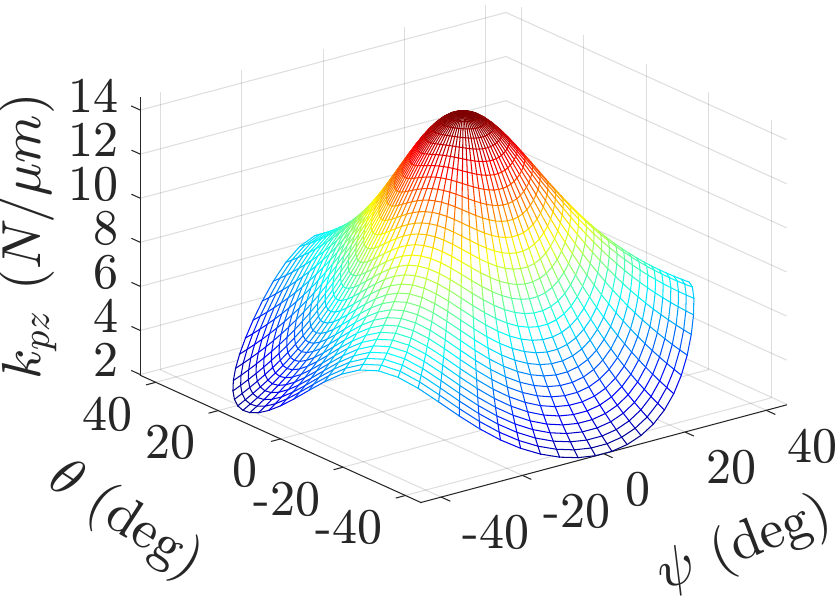}
		\caption{}\label{fig:kpz_z3}
	\end{subfigure}\hfill
	\caption{The z-axis axial stiffness distribution over rotational workspace.  (a) Sprint Z3. (b)  A3 head}
	\label{fig:kpz}
\end{figure}

\begin{figure}[!htb]
	\centering
	\begin{subfigure}{0.5\columnwidth}
		\centering
		\includegraphics[width=\textwidth]{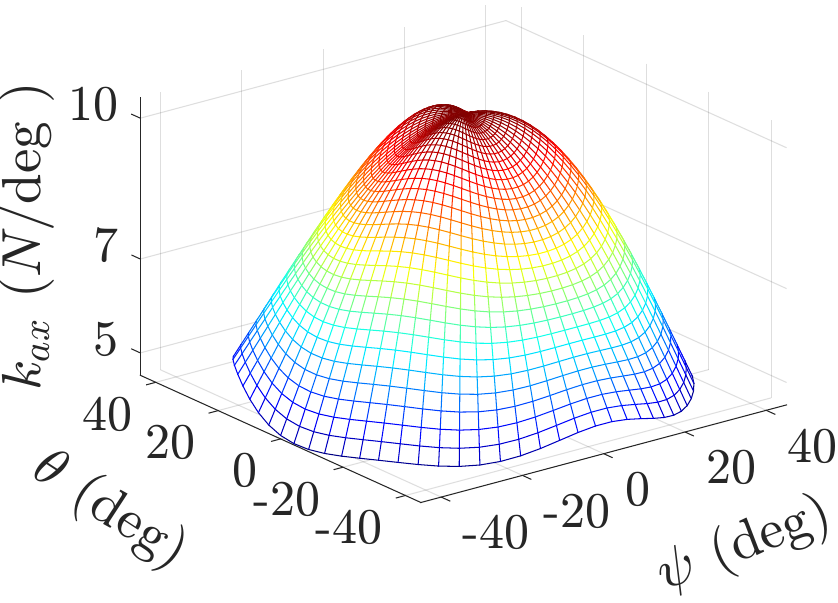}
		\caption{}\label{fig:kax_A3}
	\end{subfigure}\hfill
	\begin{subfigure}{0.5\columnwidth}
		\centering
		\includegraphics[width=\textwidth]{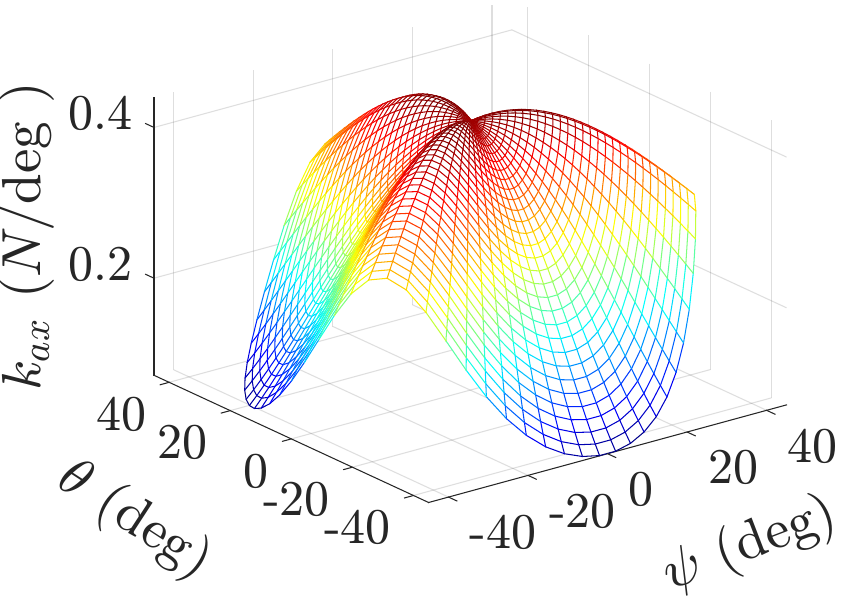}
		\caption{}\label{fig:kax_z3}
	\end{subfigure}\hfill
	\caption{The x-axis torsional stiffness distribution over rotational workspace.  (a) Sprint Z3. (b)  A3 head}
	\label{fig:kax}
\end{figure}

\begin{figure}[!htb]
	\centering
	\begin{subfigure}{0.5\columnwidth}
		\centering
		\includegraphics[width=\textwidth]{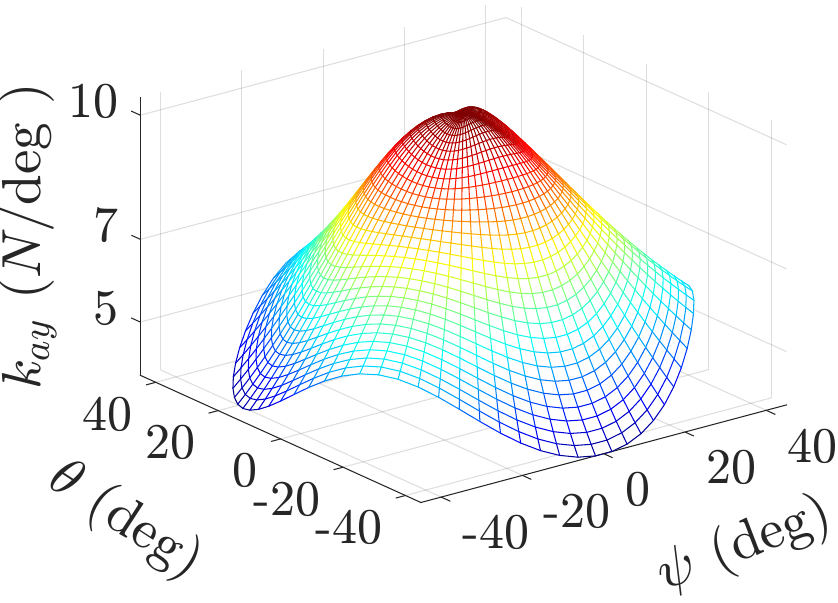}
		\caption{}\label{fig:kay_A3}
	\end{subfigure}\hfill
	\begin{subfigure}{0.5\columnwidth}
		\centering
		\includegraphics[width=\textwidth]{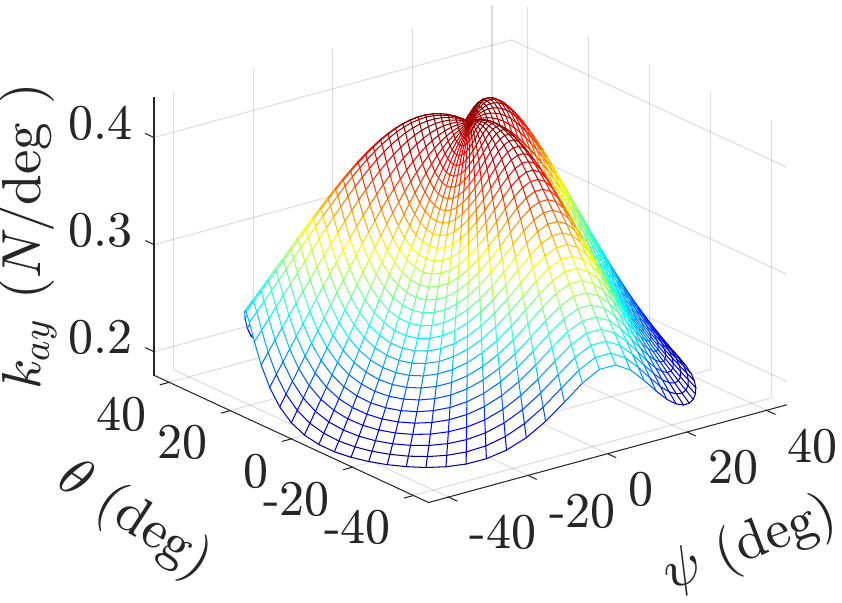}
		\caption{}\label{fig:kay_z3}
	\end{subfigure}\hfill
	\caption{The y-axis torsional stiffness distribution over rotational workspace.  (a) Sprint Z3. (b)  A3 head}
	\label{fig:kay}
\end{figure}

\begin{figure}[!htb]
	\centering
	\begin{subfigure}{0.5\columnwidth}
		\centering
		\includegraphics[width=\textwidth]{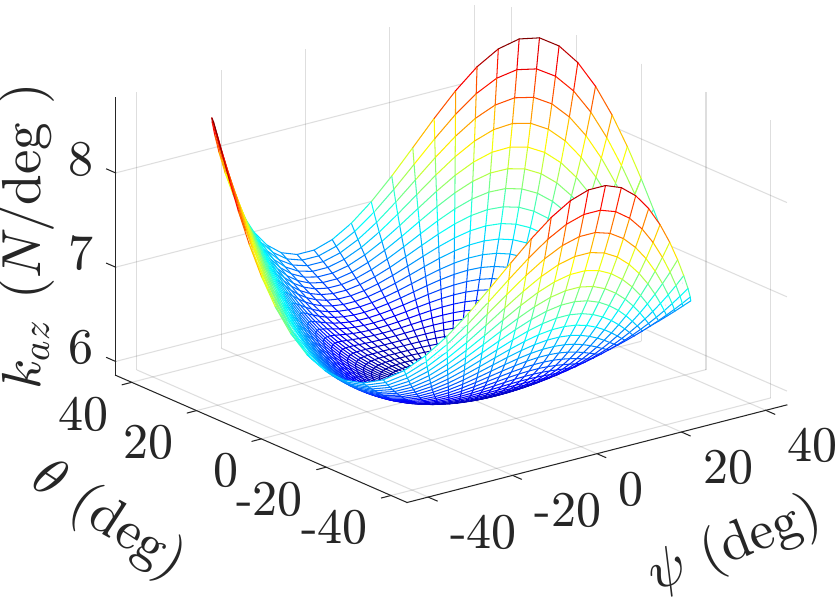}
		\caption{}\label{fig:kaz_A3}
	\end{subfigure}\hfill
	\begin{subfigure}{0.5\columnwidth}
		\centering
		\includegraphics[width=\textwidth]{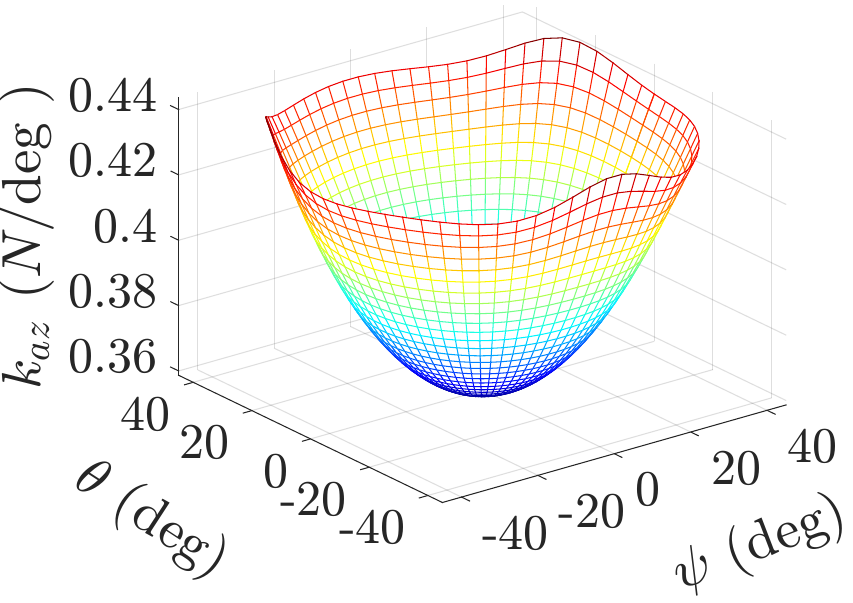}
		\caption{}\label{fig:kaz_z3}
	\end{subfigure}\hfill
	\caption{The z-axis torsional stiffness distribution over rotational workspace.  (a) Sprint Z3. (b)  A3 head}
	\label{fig:kaz}
\end{figure}

\begin{figure}[!htb]
	\centering
	\begin{subfigure}{0.5\columnwidth}
		\centering
		\includegraphics[width=\textwidth]{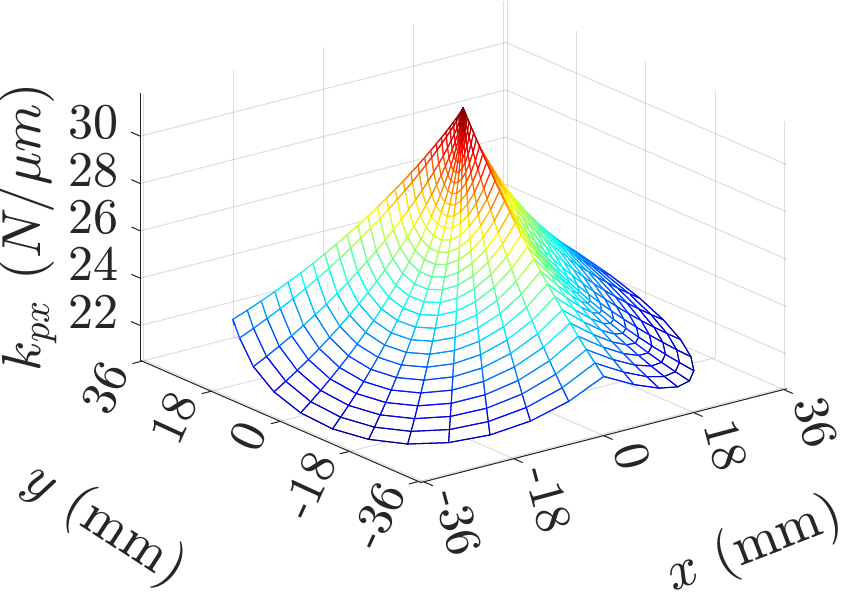}
		\caption{}\label{fig:kpx_A3_param}
	\end{subfigure}\hfill
	\begin{subfigure}{0.5\columnwidth}
		\centering
		\includegraphics[width=\textwidth]{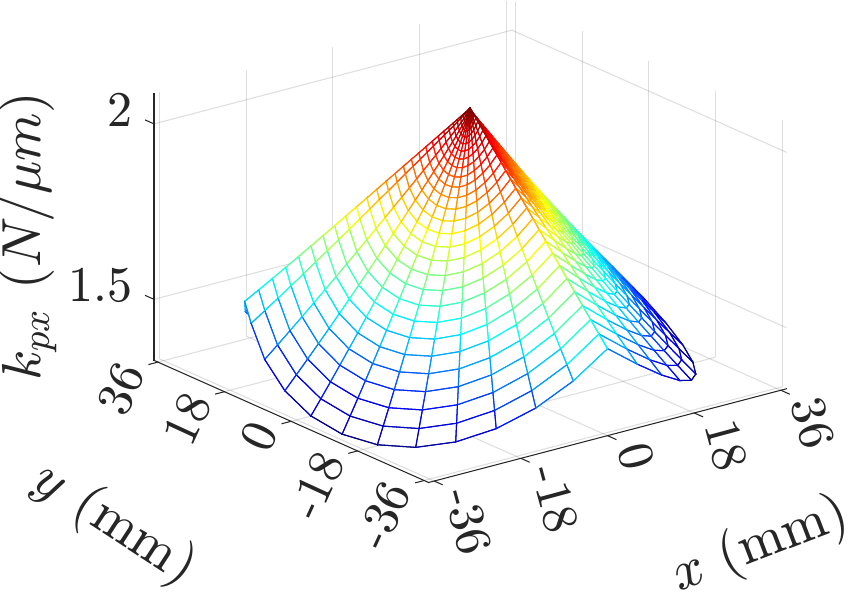}
		\caption{}\label{fig:kpx_z3_param}
	\end{subfigure}\hfill
	\caption{The x-axis axial stiffness distribution over parasitic space.  (a) Sprint Z3. (b)  A3 head}
	\label{fig:kpx_param}
\end{figure}

\begin{figure}[!htb]
	\centering
	\begin{subfigure}{0.5\columnwidth}
		\centering
		\includegraphics[width=\textwidth]{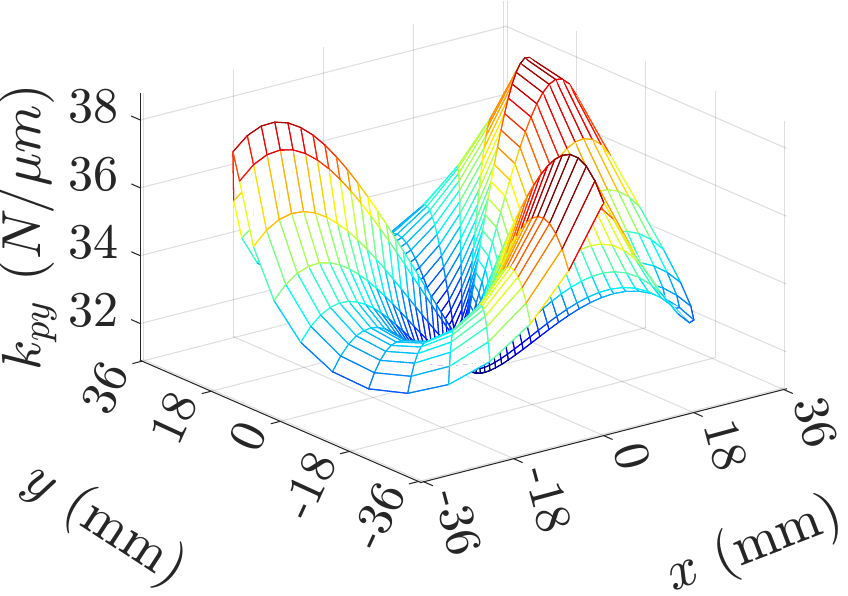}
		\caption{}\label{fig:kpy_A3_param}
	\end{subfigure}\hfill
	\begin{subfigure}{0.5\columnwidth}
		\centering
		\includegraphics[width=\textwidth]{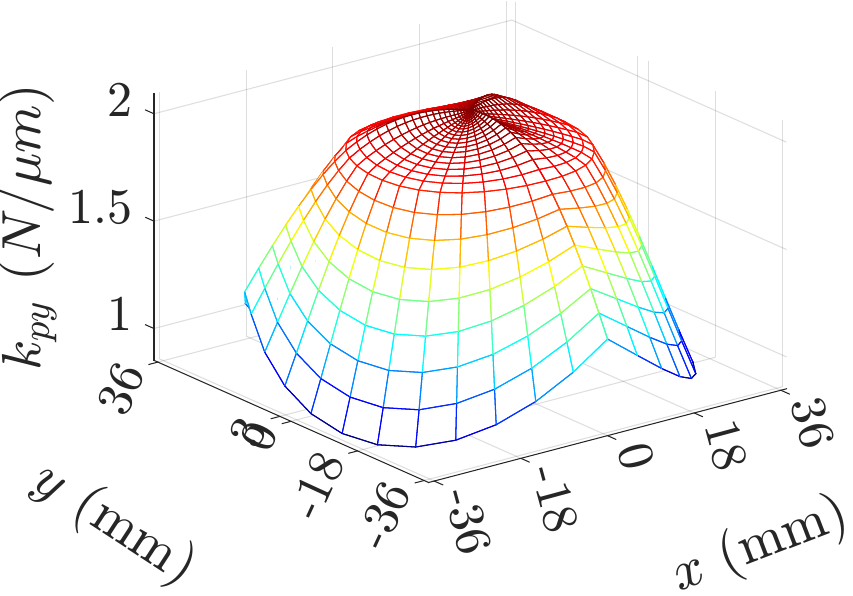}
		\caption{}\label{fig:kpy_z3_param}
	\end{subfigure}\hfill
	\caption{The y-axis axial stiffness distribution over parasitic space.  (a) Sprint Z3. (b)  A3 head}
	\label{fig:kpy_param}
\end{figure}

\begin{figure}[!htb]
	\centering
	\begin{subfigure}{0.5\columnwidth}
		\centering
		\includegraphics[width=\textwidth]{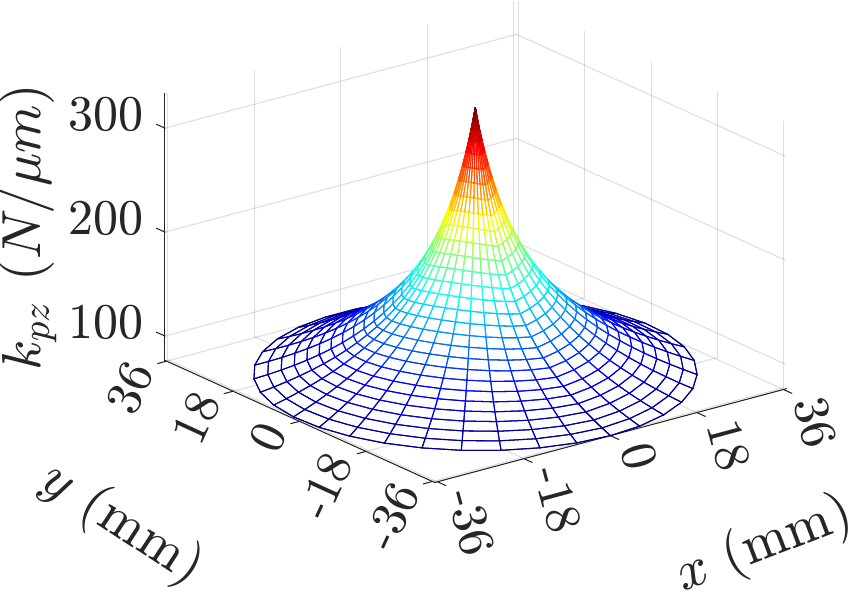}
		\caption{}\label{fig:kpz_A3_param}
	\end{subfigure}\hfill
	\begin{subfigure}{0.5\columnwidth}
		\centering
		\includegraphics[width=\textwidth]{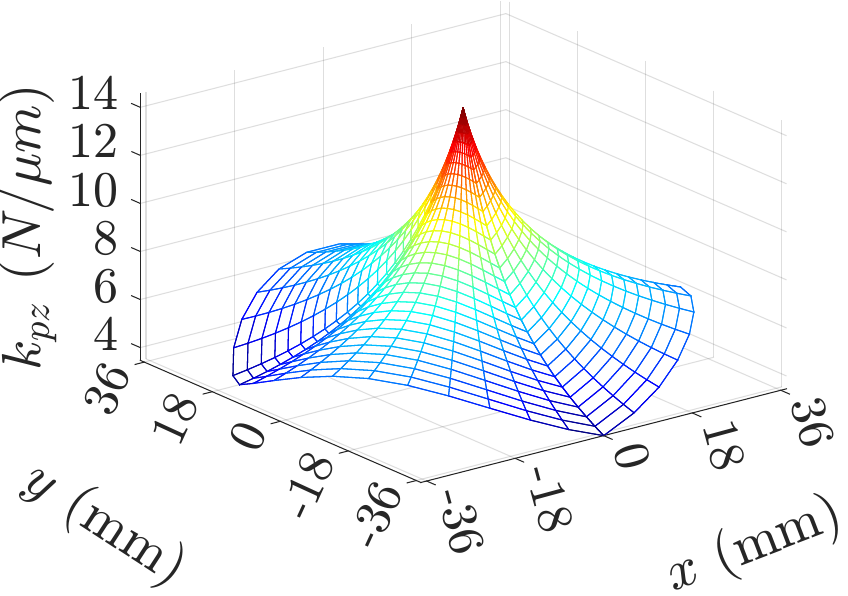}
		\caption{}\label{fig:kpz_z3_param}
	\end{subfigure}\hfill
	\caption{The z-axis axial stiffness distribution over parasitic space.  (a) Sprint Z3. (b)  A3 head}
	\label{fig:kpz_param}
\end{figure}

\begin{figure}[!htb]
	\centering
	\begin{subfigure}{0.5\columnwidth}
		\centering
		\includegraphics[width=\textwidth]{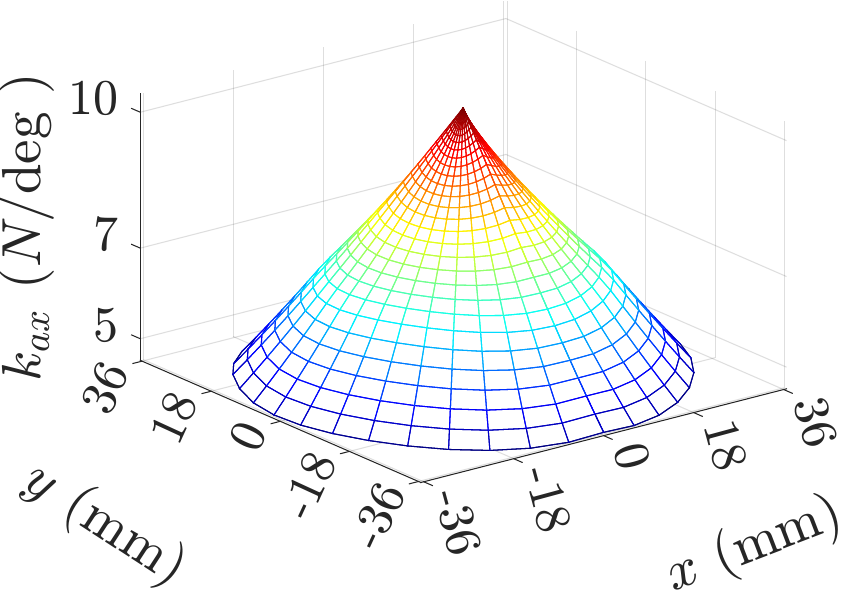}
		\caption{}\label{fig:kax_A3_param}
	\end{subfigure}\hfill
	\begin{subfigure}{0.5\columnwidth}
		\centering
		\includegraphics[width=\textwidth]{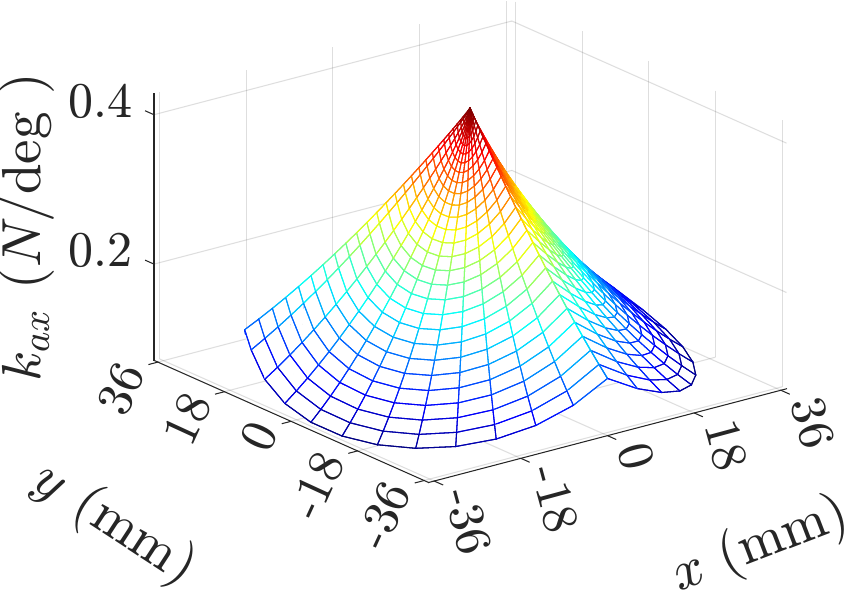}
		\caption{}\label{fig:kax_z3_param}
	\end{subfigure}\hfill
	\caption{The x-axis torsional stiffness distribution over parasitic space.  (a) Sprint Z3. (b)  A3 head}
	\label{fig:kax_param}
\end{figure}

\begin{figure}[!htb]
	\centering
	\begin{subfigure}{0.5\columnwidth}
		\centering
		\includegraphics[width=\textwidth]{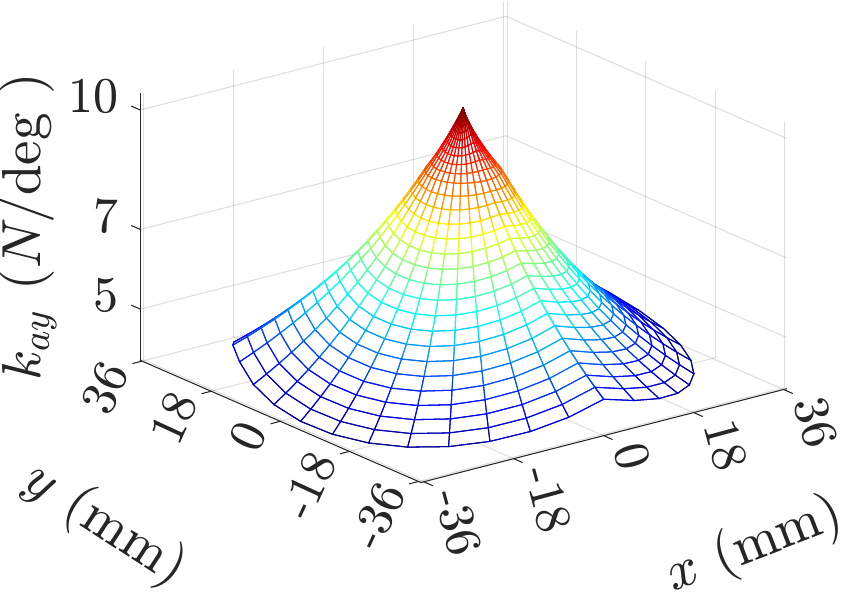}
		\caption{}\label{fig:kay_A3_param}
	\end{subfigure}\hfill
	\begin{subfigure}{0.5\columnwidth}
		\centering
		\includegraphics[width=\textwidth]{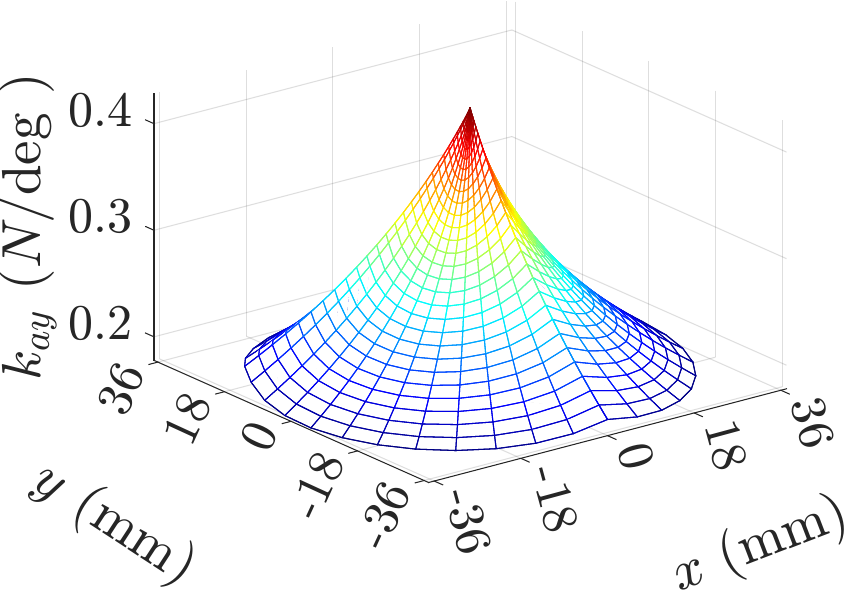}
		\caption{}\label{fig:kay_z3_param}
	\end{subfigure}\hfill
	\caption{The y-axis torsional stiffness distribution over parasitic space.  (a) Sprint Z3. (b)  A3 head}
	\label{fig:kay_param}
\end{figure}

\begin{figure}[!htb]
	\centering
	\begin{subfigure}{0.5\columnwidth}
		\centering
		\includegraphics[width=\textwidth]{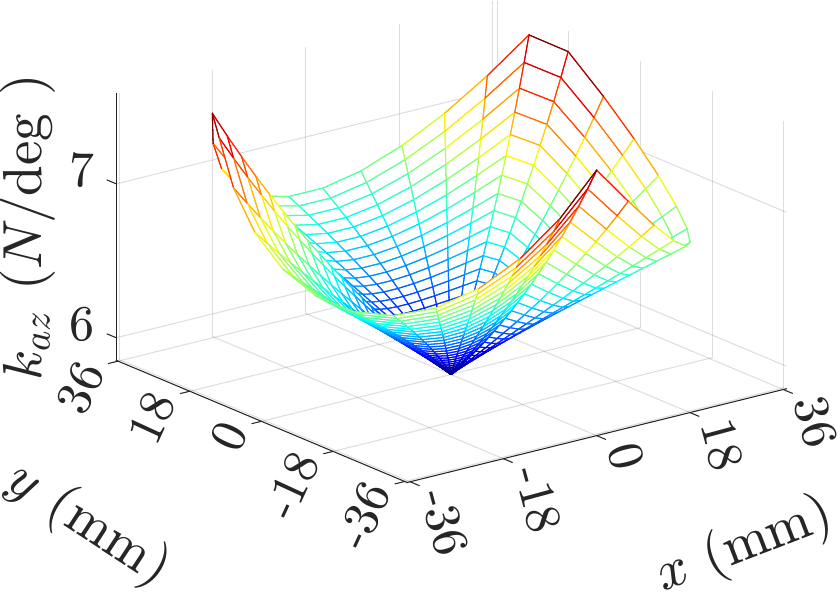}
		\caption{}\label{fig:kaz_A3_param}
	\end{subfigure}\hfill
	\begin{subfigure}{0.5\columnwidth}
		\centering
		\includegraphics[width=\textwidth]{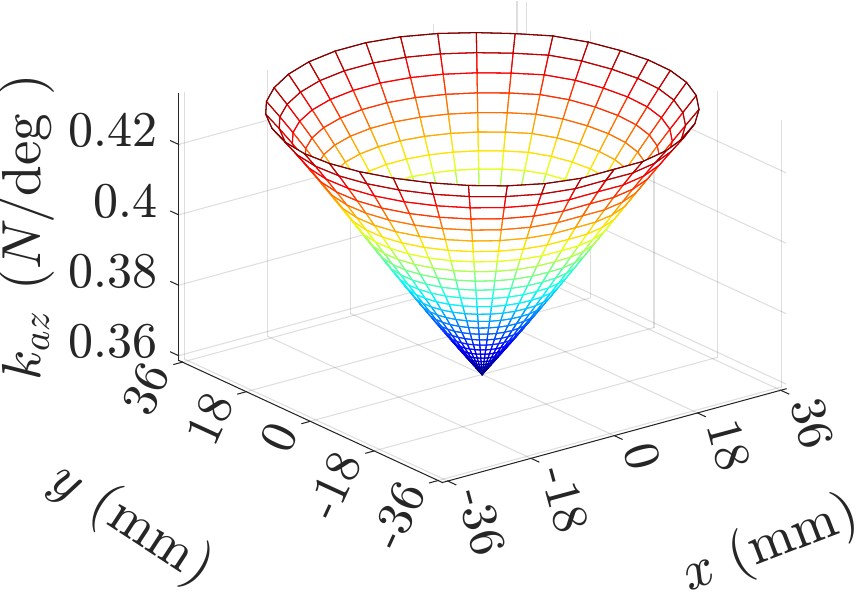}
		\caption{}\label{fig:kaz_z3_param}
	\end{subfigure}\hfill
	\caption{The z-axis torsional stiffness distribution over parasitic space.  (a) Sprint Z3. (b)  A3 head}
	\label{fig:kaz_param}
\end{figure}

Firstly, the distribution of parasitic motion across the rotational workspace is assessed for both manipulators. This analysis, depicted in Figs. \ref{fig:x_pa}, \ref{fig:y_pa}, and \ref{fig:z_pa}, is performed using the relationship defined in Eq. (\ref{eq:7}). The simulation results reveal that both manipulators exhibit identical amplitudes of parasitic motion in all directions. Based on the condition number of dimensionally homogeneous Jacobian, the condition number distribution and workspace capability are evaluated. The results displayed in Fig. \ref{fig:cond_no} and Fig. \ref{fig:workspace} shows the condition numbers and workspace of the two machines, respectively, revealing the superiority of the Sprint Z3. The workspace of the A3 head significantly reduces as the height decreases, whereas the workspace of the Sprint Z3 remains unchanged. This difference is due to the fact that $l_i$ is telescopic link in the A3 head and fixed-length link in the Sprint Z3. Then, the axial and torsional stiffness across the rotation workspace ($\theta$ and $\psi$ ) are evaluated as shown from Fig. \ref{fig:kpx} to Fig. \ref{fig:kaz}. Moreover, the stiffness distribution result across the parasitic space is shown from Fig. \ref{fig:kpx_param} to Fig. \ref{fig:kaz_param}.

While comparing the stiffness distribution across the rotational and parasitic space within the same mechanism, both devices show better stiffness in the orientation workspace. Observing the results in Figs. \ref{fig:kpx_param} to \ref{fig:kaz_param}, the stiffness profile narrows like a cone, while the stiffness profile across the rotational workspace, as shown from Fig. \ref{fig:kpx} to Fig. \ref{fig:kaz}, covers a wider range of $\psi$ and $\theta$, giving it a parabolic shape. This suggests that it is not possible to conclude the manipulators' performance by only observing the stiffness profile over the rotational workspace. Therefore, evaluating both the axial and torsional stiffness in both spaces is crucial.

The results of this work reveal critical insights into the performance of the two machines. The parasitic motion amplitude of the two mechanisms remains identical in any arbitrary configuration. The condition number also deteriorates as the manipulators approach singularity, although the Sprint Z3 machine still displays a higher peak value. We can also observe that both manipulators exhibit a peak in stiffness at the home position, with the Sprint Z3 demonstrating notably higher stiffness values, suggesting its superior ability to resist axial and torsional loads. The A3 head maintains a more uniform stiffness across its rotational workspace, potentially beneficial for applications requiring consistent performance in various orientations.

\section{Conclusion}

This paper presented a comprehensive analysis and comparison of the Sprint Z3 and A3 machining heads. Parasitic motion, condition number, workspace, and stiffness characteristics were used to measure and compare their performance. The parasitic motion is formulated at the velocity level and is integrated to obtain the pose information. The dimensionally homogeneous Jacobian is used to evaluate the condition number distribution over the orientation workspace. Then, the stiffness capability of the machines across the orientational and parasitic motion space is evaluated. The stiffness performance in relation to the parasitic motion is assessed for the first time with the aid of a constraint projection matrix. Despite operating under identical conditions, parameters and exhibiting similar magnitudes of parasitic motion, our results reveal significant differences in overall performance. Notably, the Sprint Z3 demonstrated superior capability in workspace, condition number and stiffness, signifying its enhanced ability to resist deflections that could potentially impair machining accuracy. This discrepancy is primarily attributed to the intrinsic mechanical design differences between the two machines, especially concerning the configuration of joints and links. The Sprint Z3’s architecture inherently offers a more robust response to external forces, pivotal for the precision required in manufacturing large components. This machine also provides a consistent workspace regardless of stroke variation. In conclusion, our study not only sheds light on the critical factors that determine the performance of PKMs in industrial settings but also provides a foundation for making informed decisions when selecting machinery for specific applications. The superior performance of the Sprint Z3, as demonstrated through our analysis, establishes it as the preferable choice for tasks where rigidity and precision are paramount.
 
\section*{Acknowledgment}

This research was supported by the Robotics Research Center of Yuyao (Grant No. KZ22308), National Natural Science Foundation of China under the Youth Program (Grant No. 509109-N72401) and the 2023 National High-level Talent Project, also within the Youth Program (Grant No. 588020-X42306/008).

\nocite{*}
\bibliographystyle{IEEEtran}
\bibliography{IEEEabrv,IEEEexample}

\begin{thebibliography}{10}
\providecommand{\url}[1]{#1}
\csname url@samestyle\endcsname
\providecommand{\newblock}{\relax}
\providecommand{\bibinfo}[2]{#2}
\providecommand{\BIBentrySTDinterwordspacing}{\spaceskip=0pt\relax}
\providecommand{\BIBentryALTinterwordstretchfactor}{4}
\providecommand{\BIBentryALTinterwordspacing}{\spaceskip=\fontdimen2\font plus
\BIBentryALTinterwordstretchfactor\fontdimen3\font minus
  \fontdimen4\font\relax}
\providecommand{\BIBforeignlanguage}[2]{{%
\expandafter\ifx\csname l@#1\endcsname\relax
\typeout{** WARNING: IEEEtran.bst: No hyphenation pattern has been}%
\typeout{** loaded for the language `#1'. Using the pattern for}%
\typeout{** the default language instead.}%
\else
\language=\csname l@#1\endcsname
\fi
#2}}
\providecommand{\BIBdecl}{\relax}
\BIBdecl

\bibitem{inbook}
T.~Tang, J.~Zhang, and M.~Ceccarelli, \emph{Static Performance Analysis of an
  Exechon-like Parallel Kinematic Machine}, 11 2017, vol. 408, pp. 831--843.

\bibitem{starrag}
S.~group, ``Ecospeed f 2035 - starrag,''
  \url{https://www.starrag.com/en-us/machine/ecospeed-f-2035/132}, april 2024.

\bibitem{pkmtricept}
YTEAM, ``Tricept/trimule,''
  \url{http://www.chnrobot.cn/en/listPro2.aspx?cateid=122&bigcateid=114}, april
  2024.

\bibitem{ZHANG2016208}
J.~Zhang, Y.~Zhao, and Y.~Jin, ``Kinetostatic-model-based stiffness analysis of
  exechon pkm,'' \emph{Robotics and Computer-Integrated Manufacturing},
  vol.~37, pp. 208 -- 220, 2016.

\bibitem{WAHL2000}
J.~WAHL, ``{Articulated tool head},'' \emph{CA Patent CA2,349,579}, vol.~1,
  no.~12, p.~12, 2000.

\bibitem{StarragG98}
O.~aerospace industry, ``Starrag group receives large order in the us -
  orizon,''
  \url{https://www.orizonaero.com/news/single/starrag-group-receives-large-order-us/},
  April 2024.

\bibitem{LI20131577}
Y.~Li, E.~Zhang, Y.~Song, and Z.~Feng, ``Stiffness modeling and analysis of a
  novel 4-dof pkm for manufacturing large components,'' \emph{Chinese Journal
  of Aeronautics}, vol.~26, no.~6, pp. 1577 -- 1585, 2013.

\bibitem{LIAN2016190}
B.~Lian, T.~Sun, Y.~Song, and X.~Wang, ``Passive and active gravity
  compensation of horizontally-mounted 3-rps parallel kinematic machine,''
  \emph{Mechanism and Machine Theory}, vol. 104, pp. 190 -- 201, 2016.

\bibitem{YU2018137}
\BIBentryALTinterwordspacing
G.~Yu, L.~Wang, J.~Wu, D.~Wang, and C.~Hu, ``{Stiffness modeling approach for a
  3-DOF parallel manipulator with consideration of nonlinear joint
  stiffness},'' \emph{Mechanism and Machine Theory}, vol. 123, pp. 137--152,
  may 2018. [Online]. Available:
  \url{https://linkinghub.elsevier.com/retrieve/pii/S0094114X17307231}
\BIBentrySTDinterwordspacing

\bibitem{0954406215586233}
Y.-Q. Zhao, J.~Zhang, L.-Y. Ruan, H.-W. Luo, and X.-L. Yu, ``A modified
  elasto-dynamic model based static stiffness evaluation for a 3-prs pkm,''
  \emph{Proceedings of the Institution of Mechanical Engineers, Part C: Journal
  of Mechanical Engineering Science}, vol. 230, no.~3, pp. 353--366, 2016.

\bibitem{Nigatu2020}
H.~Nigatu and Y.~Yihun, ``{Algebraic Insight on the Concomitant Motion of 3RPS
  and 3PRS PKMs},'' in \emph{Mechanisms and Machine Science}, 2020, vol.~83,
  pp. 242--252.

\bibitem{Nigatu2021}
H.~Nigatu, Y.~{Ho Choi}, and D.~Kim, ``{On the Structural Constraint and Motion
  of 3-PRS Parallel Kinematic Machines},'' in \emph{Volume 8A: 45th Mechanisms
  and Robotics Conference (MR)}, vol. 8A-2021.\hskip 1em plus 0.5em minus
  0.4em\relax American Society of Mechanical Engineers, aug 2021.

\bibitem{Nigatu2021_opt}
H.~Nigatu and D.~Kim, ``{Optimization of 3-DoF Manipulators' Parasitic Motion
  with the Instantaneous Restriction Space-Based Analytic Coupling Relation},''
  \emph{Applied Sciences}, vol.~11, no.~10, p. 4690, may 2021.

\bibitem{Chen2014}
X.~Chen, X.~J. Liu, F.~G. Xie, and T.~Sun, ``{A comparison study on
  motion/force transmissibility of two typical 3-DOF parallel manipulators: The
  sprint Z3 and A3 tool heads},'' \emph{International Journal of Advanced
  Robotic Systems}, vol.~11, no.~1, pp. 1--10, 2014.

\bibitem{article}
D.~Kim and W.~Chung, ``Analytic formulation of reciprocal screws and its
  application to nonredundant robot manipulators,'' \emph{Journal of Mechanical
  Design - J MECH DESIGN}, vol. 125, 03 2003.

\bibitem{Nigatu2023}
H.~Nigatu and D.~Kim, ``{Workspace optimization of 1T2R parallel manipulators
  with a dimensionally homogeneous constraint-embedded Jacobian},''
  \emph{Mechanism and Machine Theory}, vol. 188, p. 105391, oct 2023.

\bibitem{Nigatu2021_mmt}
H.~Nigatu, Y.~H. Choi, and D.~Kim, ``{Analysis of parasitic motion with the
  constraint embedded Jacobian for a 3-PRS parallel manipulator},''
  \emph{Mechanism and Machine Theory}, vol. 164, p. 104409, oct 2021.

\bibitem{Nigatu2023ICCAS}
H.~Nigatu and D.~Kim, ``{Dimensionally Homogeneous Jacobian using Extended
  Selection Matrix for Performance Evaluation and Optimization of Parallel
  Manipulators},'' in \emph{2023 The 23rd International Conference on Control,
  Automation and Systems}, no. Iccas, 2023.

\bibitem{Merlet2007}
J.~P. Merlet, ``{Jacobian, Manipulability, Condition Number and Accuracy of
  Parallel Robots},'' in \emph{Robotics Research}.\hskip 1em plus 0.5em minus
  0.4em\relax Berlin, Heidelberg: Springer Berlin Heidelberg, 2007, vol.~28,
  pp. 175--184.

\bibitem{Pond2006}
G.~Pond and J.~A. Carretero, ``{Formulating Jacobian matrices for the dexterity
  analysis of parallel manipulators},'' \emph{Mechanism and Machine Theory},
  vol.~41, no.~12, pp. 1505--1519, dec 2006.

\bibitem{Ruiz2016}
A.~Ruiz, F.~J. Campa, C.~Rold{\'{a}}n-Paraponiaris, O.~Altuzarra, and C.~Pinto,
  ``{Experimental validation of the kinematic design of 3-PRS compliant
  parallel mechanisms},'' \emph{Mechatronics}, vol.~39, pp. 77--88, 2016.

\bibitem{Huang2013}
Z.~Huang, Q.~Li, and H.~Ding, \emph{Mobility Analysis Part-1}.\hskip 1em plus
  0.5em minus 0.4em\relax Dordrecht: Springer Netherlands, 2013, pp. 47--69.

\bibitem{Carretero2000}
J.~A. Carretero, R.~P. Podhorodeski, M.~A. Nahon, and C.~M. Gosselin,
  ``{Kinematic analysis and optimization of a new three degree-of-freedom
  spatial parallel manipulator},'' \emph{Journal of Mechanical Design,
  Transactions of the ASME}, vol. 122, no.~1, pp. 17--24, 2000.
\renewcommand{\BIBentryALTinterwordstretchfactor}{4}

\end{thebibliography}

\end{document}